\documentclass[lettersize,journal]{IEEEtran}
\usepackage{amsmath,amsfonts}
\usepackage{algorithmic}
\usepackage{algorithm}
\usepackage{array}
\usepackage[caption=false,font=normalsize,labelfont=sf,textfont=sf]{subfig}
\usepackage{textcomp}
\usepackage{stfloats}
\usepackage{url}
\usepackage{verbatim}
\usepackage{graphicx}
\usepackage{epstopdf}
\usepackage{cite}
\usepackage{color}
\usepackage{xcolor}
\usepackage{enumerate}
\usepackage{diagbox}
\usepackage{booktabs}
\usepackage{multirow}
\usepackage{tabularx}
\usepackage{makecell}
\usepackage{amssymb}

\usepackage{xspace}
\newcommand*{\eg}{\textit{e.g.}\@\xspace}
\newcommand*{\ie}{\textit{i.e.}\@\xspace}
\newcommand*{\core}{\text{core}}
\usepackage{changes}
\begin{document}
\title{CoreMark: Toward Robust and Universal Text Watermarking Technique}
\author{Jiale Meng, ~Yiming Li, ~Zhe-Ming Lu,~\IEEEmembership{Senior Member,~IEEE}, ~Zewei He, ~Hao Luo, ~Tianwei Zhang
\thanks{Jiale Meng, Zhe-Ming Lu, Zewei He, and Hao Luo are with the School of Aeronautics and Astronautics, Zhejiang University, 310027 Hangzhou, P. R. China (e-mail: \{mengjiale, zheminglu, zeweihe, luohao\}@zju.edu.cn).}
\thanks{Yiming Li and Tianwei Zhang are with the College of Computing and Data Science, Nanyang Technological University, Singapore (e-mail: liyiming.tech@gmail.com, tianwei.zhang@ntu.edu.sg).}
\thanks{Corresponding author: Zhe-Ming Lu (zheminglu@zju.edu.cn)}
}

\markboth{PREPRINT}%
{PREPRINT}

\maketitle

\begin{abstract}
Text watermarking schemes have gained considerable attention in recent years, yet still face critical challenges in achieving simultaneous robustness, generalizability, and imperceptibility. 
This paper introduces a new embedding paradigm, termed \textsc{core}, which comprises several consecutively aligned black pixel segments. Its key innovation lies in its inherent noise resistance during transmission and broad applicability across languages and fonts.
Based on the \textsc{core}, we present a text watermarking framework named CoreMark. 
Specifically, CoreMark first dynamically extracts \textsc{core}s from characters. Then, the characters with stronger robustness are selected according to the lengths of \textsc{core}s. By modifying the size of the \textsc{core}, the hidden data is embedded into the selected characters without causing significant visual distortions.
Moreover, a general plug-and-play embedding strength modulator is proposed, which can adaptively enhance the robustness for small font sizes by adjusting the embedding strength according to the font size.
Experimental evaluation indicates that CoreMark demonstrates outstanding generalizability across multiple languages and fonts. Compared to existing methods, CoreMark achieves significant improvements in resisting screenshot, print-scan, and print-camera attacks, while maintaining satisfactory imperceptibility.

\end{abstract}


\section{Introduction}
\label{introduction}

\IEEEPARstart{M}{any} important textual data, such as medical records, personnel files, and financial documents, have been digitized and stored. However, with the widespread dissemination of information in the digital era, textual images are increasingly vulnerable to distribution via various channels, including photography, screenshots, and scanning. Consequently, ensuring the traceability of data has become an urgent and pressing demand \cite{wei2024pointncbw, li2023black, shu2015privacy}. However, most existing watermarking methods are designed for color images \cite{Gao2025Screen, hu2025mask, Xiao2024Client}. Due to the inherently simple structure of text images and the absence of complex color and texture features, these methods cannot be directly applied to document images. Therefore, the development of watermarking techniques specifically tailored for text images remains a crucial research direction. 

Text watermarking aims to embed specific identifier information without compromising text readability, offering a viable solution for text traceability. Arguably, an effective text watermarking technique should fulfill three key requirements: robustness, generalizability, and invisibility. These can be defined as follows: (i) Robustness: During document transmission and management, text images commonly undergo operations such as print-scan, print-camera capture, and screenshots. Therefore, text watermarking methods must ensure that watermarks remain extractable after such cross-media transmission processes.  (ii) Generalizability: The embedding and extraction algorithms must be applicable across various languages and fonts. (iii) Invisibility: After character encoding, the watermark should remain imperceptible to human observers, maintaining the visual integrity of the original text.

\begin{figure}[t]
\centering
\includegraphics[width=0.47\textwidth]{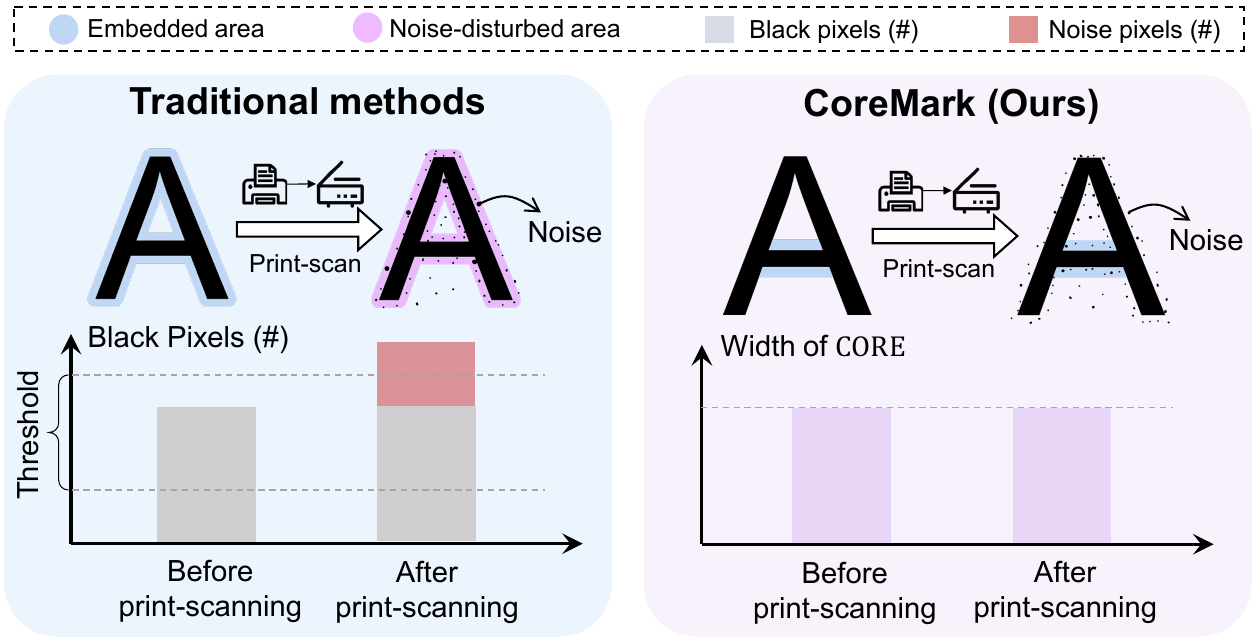}
\caption{\textbf{Comparison between traditional image-based methods and CoreMark (ours) under print-scan distortion.} Traditional image-based methods are easily disturbed by noise, which can cause the black pixel count to exceed the threshold and lead to extraction failure. In contrast, our CoreMark maintains stable embedding features under noise.}
\label{fig:intro}
\vspace{-0.5em}
\end{figure}

Current text watermarking techniques primarily include image-based methods \cite{wu2004data, tan2019print, yang2023language} and font-based methods \cite{yang2023autostegafont, qi2019robust, xiao2018fontcode}. Image-based methods treat characters as images and embed information by modifying pixels to alter their statistical characteristics. Font-based methods establish a character codebook through manual design or deep learning and embed data by substituting original characters with variants from the codebook. However, neither approach addresses the three critical requirements simultaneously. Specifically, although existing image-based text watermarking techniques exhibit strong generalizability because they operate at the pixel level and are inherently independent of specific languages or font styles. However, they are fragile in terms of robustness \cite{wu2004data}, primarily due to the widespread noise introduced during transmission \cite{solanki2006print, yu2005print}, as shown in Figure~\ref{fig:intro}. While font-based methods are generally less susceptible to noise compared to image-based approaches, they suffer from limited generalizability, as it is virtually impossible for codebooks to encompass all characters across diverse fonts and languages \cite{yang2023autostegafont, yang2023language}. Beyond the above limitations, we revisit existing image-based and font-based text watermarking methods and identify additional potential limitations. Specifically, we reveal that image-based methods often exhibit a trade-off between robustness and invisibility, and may occasionally fail to embed watermarks successfully. Font-based methods that rely on edge features are susceptible to binarization. Consequently, text image protection remains a significant challenge. Arguably, the difficulty stems from the interdependence of these three requirements: robustness, generalizability, and invisibility, makeing them difficult to balance effectively. An intriguing and important question arises: \textit{Is it possible to design a text image watermarking scheme that is simultaneously robust, generalizable, and invisible?}

The answer to the above-mentioned question is positive, although its solution is non-trivial. Motivated by the fact that existing image-based text watermarking techniques generally have a high generalizability, this paper aims to design a new image-based watermarking method that enhances robustness and imperceptibility while maintaining such generalizability.
As shown in Figure~\ref{fig:intro}, traditional approaches embed information across broad regions, rendering them vulnerable to noise interference. Typically, noise introduced during transmission exhibits two key characteristics: \textbf{1)} it is widely distributed yet spatially non-uniform, and \textbf{2)} it appears in a discrete manner. Building on this observation, we focus on the local structural features of characters, \ie, specific sub-regions within a single character, rather than relying on global embedding. This allows us to reduce embedding regions and mitigate the impact of noise. To further mitigate interference from noise surrounding the embedding area, we propose a novel embedding paradigm for character embedding, termed \textsc{core}, which comprises several consecutively aligned black pixel segments. This paradigm is inherently immune to such distortions while embedding information through minimal pixel modifications, achieving an optimal balance between robustness and imperceptibility. Furthermore, we design a \textsc{core} extraction module that adaptively identifies optimal \textsc{core}s from diverse characters, ensuring superior generalizability.
Specifically, \textsc{core} provides three advantageous properties: \textbf{1)} the inherent continuity of \textsc{core} structures effectively counters noise interference, which typically manifests as randomly distributed discrete pixels; \textbf{2)} \textsc{core} occupies minimal image area, enabling targeted local modifications during embedding and extraction processes, thereby substantially reducing noise susceptibility; \textbf{3)} \textsc{core} enables data embedding by minimal pixel modifications, preserving high visual quality, thus with a high imperceptibility.

Based on the \textsc{core} concept, we propose an image-based text watermarking method (dubbed `CoreMark'), which consists of two main stages: watermark embedding and extraction. Specifically, during the watermark embedding stage, information is embedded into characters by modifying pixels to adjust the size of \textsc{core}. Unlike edge features that rely on contour details, the size of \textsc{core} is a structural attribute that remains preserved after binarization. In the extraction stage, an adaptive threshold is first computed, and the embedded information is then recovered by comparing the width of each character’s \textsc{core} against this threshold. In particular, we take the human perceptual factor into account by referring to the neighboring pixels of \textsc{core} to ensure the structural consistency of characters before and after embedding, thereby guaranteeing invisibility. 
Besides the improvements mentioned above, we observe that robustness declines significantly as font size decreases \cite{xiao2018fontcode, tan2019print}. To enhance robustness, traditional methods typically employ a fixed repetition strategy for embedding. However, this approach may offer little to no additional benefit for larger characters that are already sufficiently robust, and can even compromise imperceptibility. To address this limitation, we propose a general plug-and-play embedding strength (ES) modulator that adaptively increases the embedding strength as the font size decreases.

In conclusion, our main contributions are fourfold.
\textbf{1)} We reveal the potential limitations of existing image-based methods and font-based approaches, along with their intrinsic mechanisms. \textbf{2)} We propose an embedding paradigm called \textsc{core}, based on which we design an image-based text watermarking method (CoreMark), to simultaneously satisfy robustness, generalizability, and invisibility. \textbf{3)} We propose a general plug-and-play embedding strength modulator to enhance robustness for characters with small font sizes by adaptively adjusting the embedding strength. \textbf{4)} Extensive experiments of various character types are conducted, including multiple fonts, languages, and font sizes, demonstrating significant performance improvements over state-of-the-art methods.
\begin{figure*}[t]
    \centering
    \begin{minipage}[c]{0.7\textwidth} 
        \centering
        \subfloat{\includegraphics[width=0.32\textwidth]{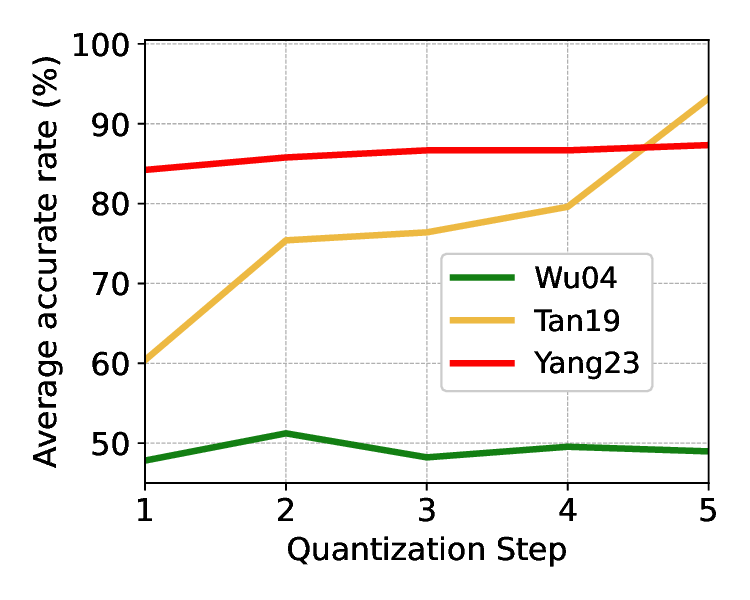}}
        \hspace{0.1em}
        \subfloat{\includegraphics[width=0.32\textwidth]{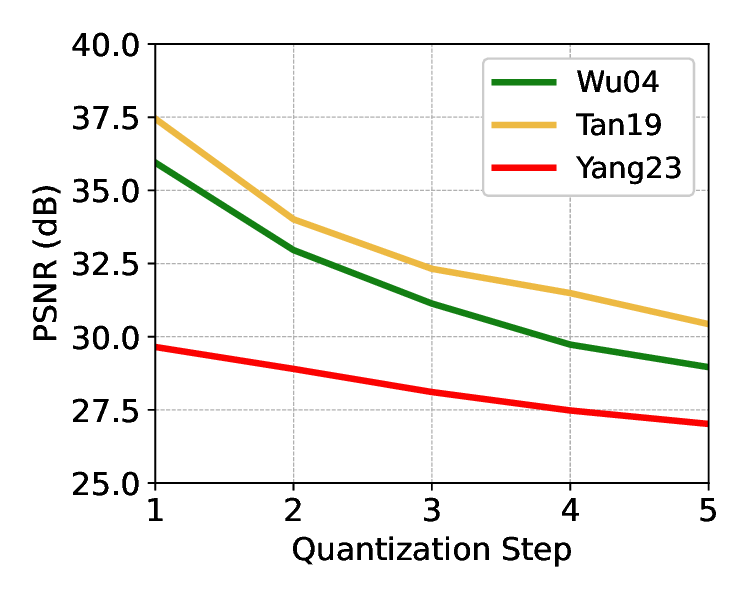}}
        \hspace{0.1em}
        \subfloat{\includegraphics[width=0.32\textwidth]{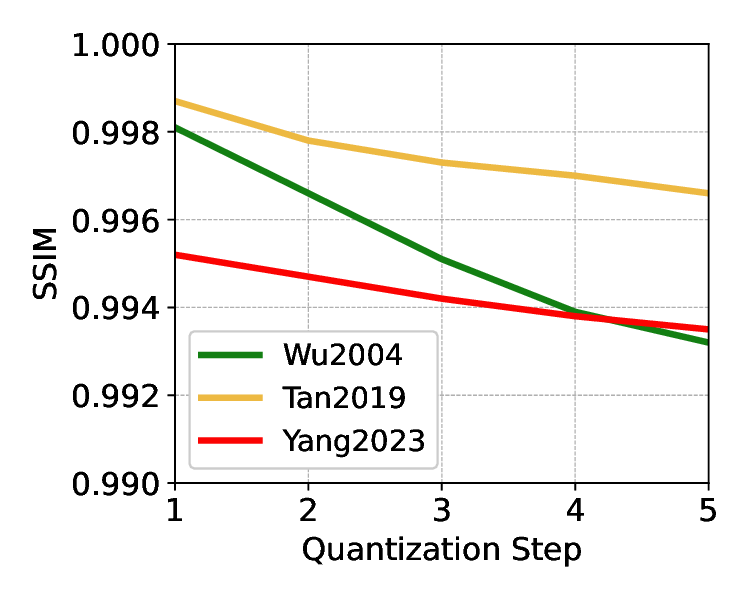}}
        \caption{Performance comparison of three image-based text watermarking methods (Wu04, Tan19, and Yang23) under varying quantization steps. Left: average extraction accuracy. Middle: PSNR. Right: SSIM.}
        \label{revisitBalance}
    \end{minipage}
    \hspace{0.4em}
    \begin{minipage}[c]{0.25\textwidth} 
        \centering
        \includegraphics[width=0.87\textwidth]{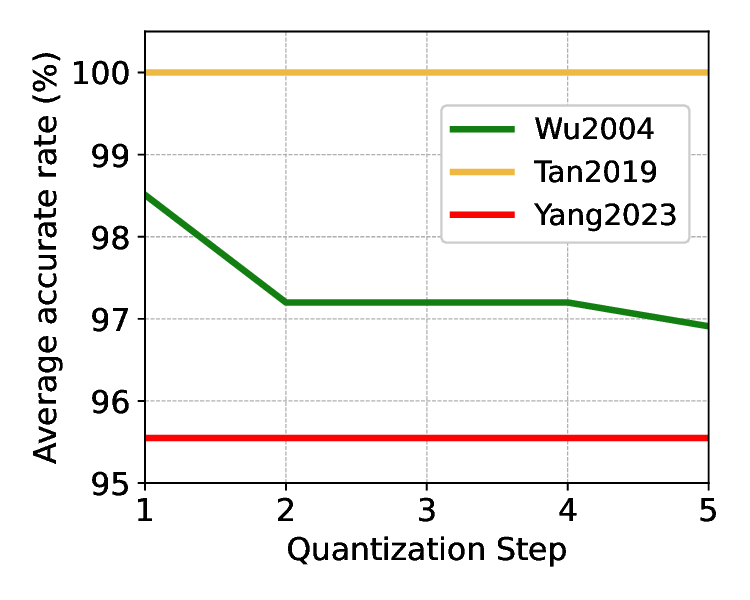}
        \caption{Robustness of methods \cite{wu2004data, tan2019print, yang2023language} without attack.}
        \label{revisit_noattack}
    \end{minipage}
    \vspace{-0.5em}
\end{figure*}

\section{related work}
\label{weakness}
Existing text watermarking techniques can be divided into four categories \cite{xie2019review}: image-based, font-based, format-based, and linguistic-based approaches. Each is discussed below.

\vspace{0.3em}
\noindent \textbf{Image-based Approach.} Image-based approaches \cite{wu2004data, qi2008document, peng2017print, peng2018print, hou2015scanned, yang2023language, tan2019print, varna2009data} handle characters as image data, hiding information by manipulating individual pixels to transform the statistical features of the characters. Wu and Liu \cite{wu2004data} firstly established a set of flipping rules and proposed a method for embedding watermarks by flipping pixels, laying an important theoretical foundation for subsequent research\cite{qi2008document, peng2017print, peng2018print, hou2015scanned}. However, the authors \cite{wu2004data} state that these approaches are only effective under high-resolution printing and scanning conditions, such as the 9600$\times$2400 dpi printer used in \cite{peng2017print}. Yang \textit{et al.} \cite{yang2023language} initially embedded data based on altering the relative center of gravity between paired characters, which shows strong language independence. Despite various efforts, a clear trade-off remains between robustness and invisibility in existing binary image watermarking methods. Methods that improve robustness typically incur visible distortions, whereas those designed for better imperceptibility often fail to withstand common degradations such as printing and scanning. This imbalance highlights the difficulty of simultaneously achieving both properties, especially in real-world degradation scenarios.

\vspace{0.3em}
\noindent \textbf{Font-based Approach.} Font-based approaches \cite{wang2022anti, yang2023autostegafont, xiao2018fontcode, qi2019robust} typically rely on a manually designed or deep learning-generated character codebook, through which data is embedded by substituting standard characters with visually similar variants. Xiao \textit{et al.} \cite{xiao2018fontcode} utilized human volunteers to identify visually similar TNR font characters, embedding information through glyph substitution and requiring 52 distinct classifiers for English letter extraction. Similarly, Qi
\textit{et al.} \cite{qi2019robust} manually deformed stroke positions to create font variants. However, Yang \textit{et al.} \cite{yang2023autostegafont} noted the limited generalizability of these methods due to the substantial manual effort required to construct a character codebook for each new font. To overcome this limitation, they proposed AutoStegaFont, an approach that automatically generates encoded fonts, representing a significant advancement in automation. Although this method eliminates human intervention, it still fails to embed information when encountering a previously unseen font.

\vspace{0.3em}
\noindent \textbf{Format-based Approach.}
The format-based approach usually includes line-shift coding \cite{brassil1995electronic, brassil1999copyright}, word-shift coding \cite{brassil1995electronic, huang2001interword, chotikakamthorn1998electronic, brassil1999copyright, zou2005formatted}, and other layout-based techniques \cite{text2007Paulo, villan2006text}. Character spacing-based watermarking approaches \cite{brassil1995electronic, brassil1999copyright} suffer from non-blind extraction requirements, necessitating access to the original document and thus significantly limiting practical deployment scenarios. Line-shift coding methods \cite{brassil1995electronic, brassil1999copyright} embed watermarks via vertical text line displacement, inherently limiting the payload capacity to fewer bits than available text lines. Huang \textit{et al.} \cite{huang2001interword} made the text line appear sine wave feature by adjusting the word space, and each text line can only embed one bit of information. Moreover, some methods \cite{wang2022anti, yang2004text} fail to embed irregularly formatted content such as equations, titles, and logos or symbols of variable size because they rely on the assumption of uniform spacing \cite{xie2019review}.

\vspace{0.3em}
\noindent \textbf{Linguistic-based Approach.}
Linguistic-based watermarking represents another established approach that embeds information by manipulating syntactic and semantic text properties \cite{meral2009natural, murphy2007syntax, chang2014practical, topkara2006hiding, meral2007syntactic, ueoka2021, 2021yang}. These methods employ syntactic transformations for watermark embedding and perform better in agglutinative languages (\eg, Korean, Arabic) because they have greater structural flexibility and inherently greater embedding capacity than English. Moreover, a defining characteristic of such approaches is their alteration of document content during watermark embedding. This poses significant limitations for sensitive documents, where semantic and syntactic modifications may compromise document integrity and value.

\section{Revisiting Existing Text Watermarking Methods}
\label{revisit}
\subsection{Preliminaries}
\vspace{0.3em}
\noindent \textbf{Threat Model.}
Our threat model comprises two entities: the defender (\ie, document owner) and the adversary. The adversary aims to illegitimately re-distribute the protected textual content through unauthorized means (\ie, print-scan, screenshot), while the defender seeks to trace the leak sources by extracting embedded watermarks from suspicious documents. We consider the most practical and stringent scenario where the adversary is aware of our method but does not have access to its specific parameters.

\vspace{0.3em}
\noindent \textbf{Representative Methods in our Evaluation.}
{\color{black}
We evaluated three image-based text watermarking methods: Wu04 \cite{wu2004data}, Tan19 \cite{tan2019print}, and Yang23 \cite{yang2023autostegafont}. 
Wu04 embeds watermarks by pixel flipping within fixed-size blocks to make black pixel counts in blocks even or odd multiples of quantization step $Q$ for bits `0' and `1', respectively. 
Tan19 modifies character boundary pixels to control the total black pixel count within each character, embedding bits based on whether the rounded ratio of black pixels to average pixels per character (scaled by $K$) is odd or even.
Yang23 manipulates relative centroid positions of adjacent characters, embedding bit 0 when the left centroid exceeds the right by threshold $R$, otherwise bit 1. The parameters $Q$, $K$, and $R$ serve as quantization steps that control embedding strength. Higher values improve robustness against attacks but reduce visual imperceptibility.}

\subsection{Challenge 1: Invisibility-Robustness Tradeoff}
\label{strategy1}
Existing image-based watermarking methods show an inherent trade-off between robustness and invisibility. Increasing the embedding strength improves robustness while degrading visual quality, thereby limiting practical deployment.

\vspace{0.3em}
\noindent \textbf{Settings.}
To quantify this trade-off, we evaluate three methods across 50 English documents rendered in Arial font. We assess imperceptibility and robustness against screenshot attacks under five different quantization steps. Specifically, Wu04 employs $Q$ values from 100 to 500 (interval: 100), Tan19 uses $K$ values from 0.05 to 0.25 (interval: 0.05), and Yang23 utilizes $R$ values from 6 to 14 (interval: 2). Robustness is measured via accuracy (ACC) defined as:
\begin{equation}
ACC = \frac{B_{\text{correct}}}{L_w} \times 100\%,
\label{ACC}
\end{equation}
where $ B_{\text{correct}} $ is the number of correctly extracted bits, and $L_w$ denotes the total number of embedded watermark bits. A higher $ ACC $ means a more robust scheme. 
Imperceptibility is evaluated using peak signal-to-noise ratio (PSNR) and structural similarity index (SSIM) \cite{wang2004image}.

\vspace{0.3em}
\noindent \textbf{Results.}
As Figure~\ref{revisitBalance} illustrates, Wu04's $ACC$ consistently hovers around 50\% regardless of the step size, equivalent to random guessing. This results from screenshot-induced resizing effects disrupting encoder-decoder synchronization due to unknown resizing factors. For Tan19, increasing the quantization step significantly improves robustness, achieving over 90\% accuracy at Step 5. However, this comes at the cost of severe imperceptibility degradation, with PSNR dropping below the commonly accepted threshold of 30 dB \cite{subhedar2020secure, setiadi2021psnr}. Yang23 consistently produces PSNR values below 30 dB due to extensive pixel modifications required to meet embedding thresholds. Meanwhile, its accuracy remains below 90\%, primarily due to embedding instability (discussed in Section~\ref{strategy2}) and susceptibility to noise. These results reveal two fundamental design principles for watermarking methods: (1) robustness is inherently linked to a method's ability to withstand scaling distortions commonly encountered in real-world transmission scenarios, such as screen capture and image resizing. Consequently, resistance to such transformations must be explicitly incorporated into method design; and (2) for image-based methods, an effective approach to mitigating the robustness-imperceptibility trade-off is to develop novel embedding paradigms that enable information embedding through modification of only a small number of pixels. This approach ensures that even with increased embedding strength, visual degradation remains acceptable.
\begin{figure}[t]
\centering
\includegraphics[width=0.43\textwidth]{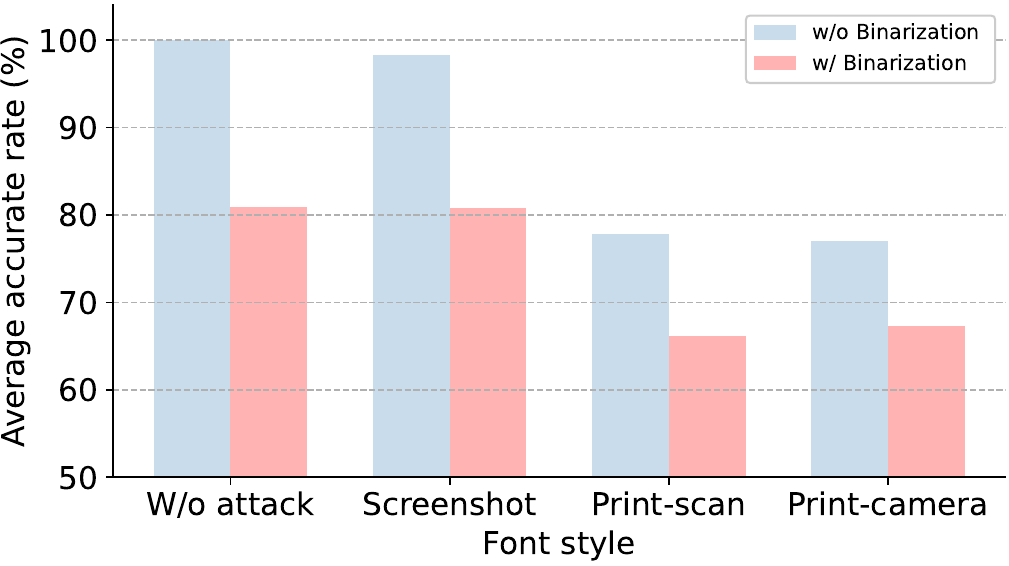}
\caption{Extraction accuracy of AutoStegaFont on 52 Calibri characters under four conditions, with and without binarization. Binarization consistently reduces performance across all scenarios, indicating its significant impact on the robustness of font-based text watermarking methods.}
\label{Binarization}
\vspace{-1em}
\end{figure}

\subsection{Challenge 2: Embedding Stability}
\label{strategy2}
We observe that certain text watermarking methods occasionally fail to embed successfully, a phenomenon we refer to as `embedding instability'.

\vspace{0.3em}
\noindent \textbf{Settings.} {\color{black}
We evaluate embedding instability using the same three methods and 50 documents from Section~\ref{strategy1}. We embed identical watermark sequences at different parameter values, extract them without attacks, and measure $ACC$. Lower $ACC$ values indicate greater embedding instability.}

\vspace{0.3em}
\noindent \textbf{Results.}
{\color{black}
As shown in Figure~\ref{revisit_noattack}, Tan19 exhibits stable embedding across all parameter values. Conversely, Wu04's instability increases as the quantization step size increases. This is due to the fact that a larger quantization step size requires more pixel modifications. When the number of modifiable pixels is limited, some characters cannot meet the embedding conditions, resulting in embedding failure. Yang23 experiences embedding failures primarily with narrow characters like `l' and `i'. These symmetric characters maintain centered centroids even after modifications, limiting the centroid difference to the embedding threshold relative to adjacent characters. As a result, they fail to satisfy the embedding criteria, leading to embedding instability.} These observations suggest two key insights: (1) ensuring all characters are suitable for embedding is challenging due to diverse character shapes and pixel distributions; and (2) developing methods that enable flexible character selection for embedding is more practical.

\subsection{Challenge 3: Resistance to Binarization}
\label{strategy3}
Binarization, a widely used preprocessing step in text image processing, often affects character edges. We observe that the font-based SOTA method AutoStegaFont \cite{yang2023autostegafont} exhibits limited robustness against binarization.

\vspace{0.3em}
\noindent \textbf{Settings.}
We evaluate $ACC$ on 52 Calibri letters under four representative conditions: no attack, screenshot, print–scan, and print–camera, both with and without binarization.

\vspace{0.3em}
\noindent \textbf{Results.}
As shown in Figure~\ref{Binarization}, AutoStegaFont achieves nearly perfect accuracy under clean and screenshot conditions without binarization. However, its accuracy drops by nearly 20\% when binarization is applied. Similar degradation is observed under print–scan and print–camera scenarios, where binarization reduces $ACC$ by over 10\%. These results highlight the vulnerability of methods that rely heavily on edge features, which are susceptible to degradation induced by binarization.

\begin{figure*}[t]
\centering
\includegraphics[width=\textwidth]{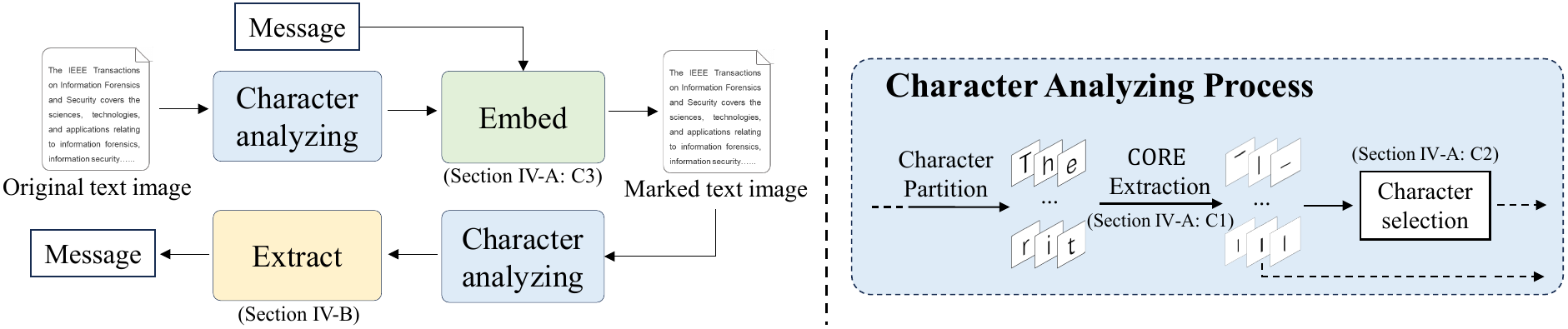}
\caption{\textbf{The overall framework of CoreMark, including embedding and extraction stages.} Embedding Stage: The original text image is segmented into individual characters, followed by \textsc{core} extraction. Robust characters are then selected for data embedding. Extraction Stage: The watermarked text image undergoes character segmentation, and the \textsc{core}s of embedded characters are extracted. Robust characters are then identified for data retrieval. Note that C1, C2, and C3 represent Component 1, Component 2, and Component 3, respectively, which correspond to the components described in Section~\ref{sec:embed}.}
\label{flowchart}
\vspace{-0.5em}
\end{figure*}

\begin{table}[t]\scriptsize
\centering
\renewcommand{\arraystretch}{1.3}
\caption{{\color{black}The Notations of Explanation.}}%
\label{notationtable}
\begin{tabular}{c|l}
\hline
Notation & Explanation \\
\hline
$\zeta_m(\mu^j_m, \rho^j_m)$ & A consecutive sequence of pixels in row $m$ with\\
& length $\mu^j_m$ and pixel value $\rho^j_m$.\\
$\zeta_m(\mu_m^{j_{\max}}, \rho_m^{j_{\max}})$ & The longest consecutive pixel sequence in row $m$ with \\
& length $\mu_m^{j_{\max}}$ and pixel value $\rho_m^{j_{\max}}$.\\
$ S_{\core} $ & \textsc{core} of each character\\
$\mathcal{I} $ &  The indices of rows belonging to $S_{\core}$. \\
$\mathcal{D}, \mathcal{P}, \mathcal{L}$ & Direction, starting point, and length of $ S_{\core} $, respectively.\\
$V_{\text{cand}}$ & Row indices of the candidate vectors.\\
$\varPhi$ & The set of row indices of candidate vectors.\\
$C_{\text{start}}$, $C_{\text{len}}$ & The starting coordinates and lengths of candidate vectors.\\
$ C$ & Set of all characters.\\
$n_{\core}$, $\tilde{n}_{\core}$ & The size of \textsc{core} of each $c_i$ before and after embedding.\\
$ T_\Delta $ & The embedding threshold.\\
$ \beta $ & The watermark embedding strength of CoreMark.\\
\hline
\end{tabular}
\end{table}

\section{The Proposed Method}
\label{section4}
To address the above challenges, we first introduce an embedding paradigm, termed \textsc{core}, based on which we introduce CoreMark, a novel watermarking method. As illustrated in Figure \ref{flowchart}, CoreMark operates in two distinct stages: watermark embedding and watermark extraction. {\color{black}The notations within this paper are shown in Table~\ref{notationtable}.}

\subsection{Watermark Embedding}
\label{sec:embed}
The embedding stage involves three components: (1) extraction of \textsc{core}s from individual characters, (2) data embedding through modification of \textsc{core} width, and (3) robustness enhancement for small font sizes via our proposed independent embedding strength (ES) modulator. 

\vspace{0.3em}
\noindent \textbf{Component 1: \textsc{core} and its Extraction Model.}
{\color{black}
A \textsc{core}, denoted as $ S_{\core}$, comprises a series of consecutive black pixels. Each \textsc{core} is determined by three fundamental parameters: direction $\mathcal{D} \in \{\text{horizontal, vertical}\}$, starting point $\mathcal{P} = \{ (x_i, y_i) \}_{i=1}^{n_{\core}}$, and length $\mathcal{L} = \{ \ell_i \}_{i=1}^{n_{\core}}$, where $n_{\core}$ is the total number of such sequences.
Specifically, $S_{\core}$ is extracted from rows if $\mathcal{D}$ is horizontal, or from columns if $\mathcal{D}$ is vertical. The starting point $\mathcal{P}$ indicates the initial pixel of each sequence, corresponding to the leftmost pixel when $\mathcal{D}$ is horizontal and the topmost pixel when $\mathcal{D}$ is vertical. The length $\mathcal{L}$ specifies the number of black pixels in each sequence.
The calculation methods for these three parameters ($\mathcal{D}, \mathcal{P}, \mathcal{L}$) are explained in detail in the following subsections.}

\begin{figure}[t]
\centering
\includegraphics[width=0.45\textwidth]{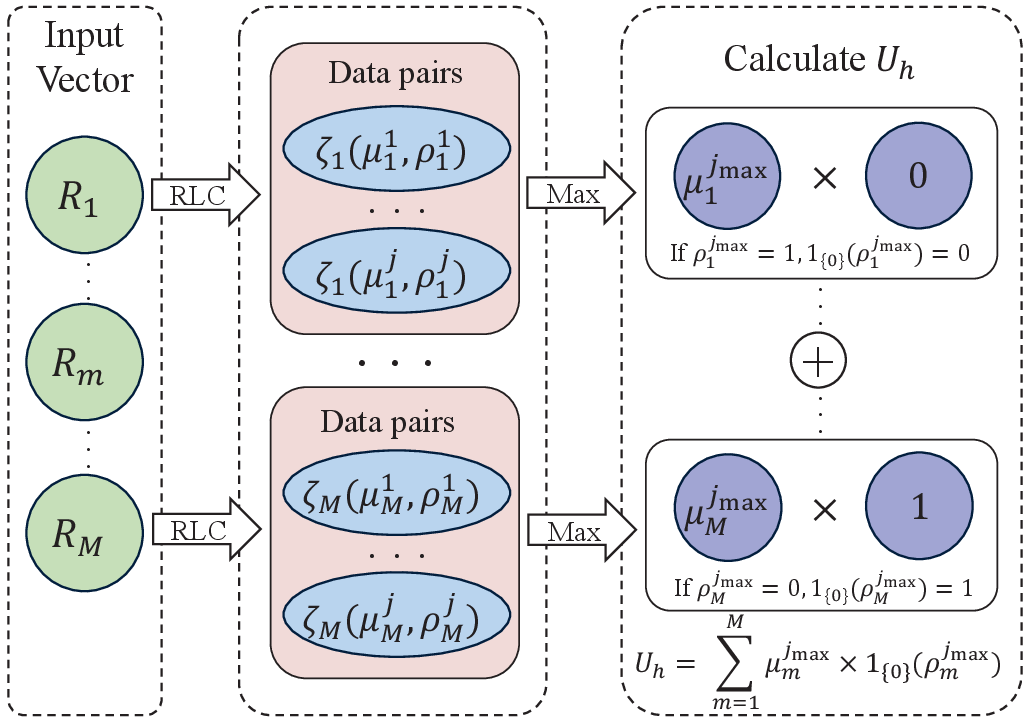}
\caption{{\color{black}Illustration of calculating $U_{h}$ in the horizontal direction.}}
\label{uh_illustrate}
\vspace{-0.5em}
\end{figure}

\noindent \textbf{(1) Determine the direction $ \mathcal{D}$.}
{\color{black}
Let the character image have dimensions $M \times N$. Taking the horizontal direction as an example, let $R_m$ denote the $m$-th row of the image, $m=1,\dots,M$.   
For each row $ R_m $, we apply Run Length Coding (RLC) \cite{thyagarajan2011still} to represent $ R_m$ as a series of data pairs $\zeta_m(\mu^j_m, \rho^j_m), j=1,\dots,J_m$, where $J_m \in \{1,\dots,N\} $. Here, $\mu^j_m \in \{1,\dots,N\}$ is the length of a consecutive pixel sequence, and $\rho^j_m \in \{0,1\}$ indicates the pixel value ($0$ for black, $1$ for white).  Let $ \mu_m^{j_{\max}} $ denote the length of the longest consecutive pixel sequence in row $m$, and $ \rho_m^{j_{\max}} $ its pixel value.
The direction $\mathcal{D}$ is determined as follows:
\begin{equation}
\mathcal{D} =
\begin{cases}
\text{horizontal}, & \text{if } \dfrac{U_h}{K_h} > \dfrac{U_v}{K_v},\\[1mm]
\text{vertical}, & \text{otherwise.}
\end{cases}
\end{equation}
where $ U_h $ is calculated by 
\begin{equation}
	U_h = \sum_{m=1}^{M}{\mu_m^{j_{\max}} \times \mathbf{1}_{\{0\}}\left(\rho_m^{j_{\max}}\right)}.
	\label{3}
\end{equation}
The indicator function $ \mathbf{1}_{\{0\}} $ is defined as:
\begin{equation}
    \mathbf{1}_{\{0\}}(x) =
    \begin{cases}
    1, & x=0,\\
    0, & x=1.
    \end{cases}
\end{equation}
$K_h= \sum_{m=1}^{M} \mathbf{1}_{\{0\}}(\rho_m^{j_{\max}})$ represents the number of rows with $ \rho_m^{j_{\max}} =0$. Analogous quantities $U_v$ and $K_v$ are computed along the vertical direction.
Figure~\ref{uh_illustrate} illustrates the calculation process of $ U_{h} $ in the horizontal direction.}

\noindent \textbf{(2) Determine the starting coordinates $\mathcal{P}$ and lengths $\mathcal{L}$.}
In general, multiple consecutive black pixel sequences exist along $\mathcal{D}$; however, not all of them qualify as elements of $ S_{\core} $. 
We illustrate the selection process in the horizontal case. We first identify candidate vectors from the RLC output. For each row, the longest black pixel run $\zeta_m(\mu_m^{j_{\max}}, \rho_m^{j_{\max}})$ with $ \rho_m^{j_{\max}}=0$ is designated as candidate vector. The set of candidate vectors is denoted by $V_{\text{cand}}$, with their row indices collected in $\varPhi$. The starting coordinates and lengths of $V_{\text{cand}}$s are defined as $C_{\text{start}} = \{ (x_t, y_t) \}_{t=1}^{\text{len}(\varPhi)}$ and $C_{\text{len}} = \{ \mu_t^{j_{\max}} \}_{t=1}^{\text{len}(\varPhi)}$.
{\color{black}
Here, $x_t=m$ is the row index, and its corresponding $y_t$ gives the starting column of the longest run, which is computed as follows.
\begin{equation}
     y_t = \sum\nolimits_{i=1}^{ {j_{\max}-1} }{\mu^i_m} + 1.
    \label{yt}
\end{equation}
}
Next, the set of candidate vector is clustered according to spatial continuity. 
\begin{algorithm}[t]
	\color{black}\caption{Method of assigning candidate vectors.}\label{alg:alg1}
	\textsl{}
	\begin{algorithmic}[1]
		\renewcommand{\algorithmicrequire}{\textbf{Input:}}
		\renewcommand{\algorithmicensure}{\textbf{Output:}}
		\REQUIRE $\varPhi$, $C_{\text{start}} = \{ (x_t, y_t) \}_{t=1}^{\text{len}(\varPhi)}$, $C_{\text{len}} = \{ \mu_t^{j_{\max}} \}_{t=1}^{\text{len}(\varPhi)}$

		\ENSURE{$\mathcal{I}$}
	
		\STATE Initiate $ C_{\text{num}}=\{ C_i^{\text{num}} \}_{i=1}^{k} $, $ C_{\text{ave}}=\{ C_i^{\text{ave}} \}_{i=1}^{k}, k \in \mathbb{Z}^+$
		\STATE \textcolor{gray}{$ // $ \textit{$ C_i^{\text{num}} $ is the number of candidate vectors in each cluster; $ C_i^{\text{ave}} $ is the average length of each cluster.}}
		\STATE boolean $b_1$, $b_2$, $b_3$
		\STATE $\hat{y} \gets y_1$, $\hat{x} \gets x_1$, $\mu_{\text{joint}} \gets \mu_{x_1}^{j_{\max}}$, $n \gets 1$, $k \gets 0$
		\FOR{each $i \in [2, \text{len}(\varPhi)]$}
		\STATE \textcolor{gray}{$ // $\textit{ Verify compliance with three rules.}}
		\STATE $b_1 \gets x_i-\hat{x}==1$
        \STATE $b_2 \gets {\rm abs}(y_i-\hat{y})\le T_c$
		\STATE $b_3 \gets {\rm abs}(y_i + \mu_{x_i}^{j_{\max}} - \hat{y} - \mu^{j_{\max}}_{\hat{x}})\le T_c $
		\IF {$ b_1 \& b_2 \& b_3$}
		\STATE \textcolor{gray}{$ // $ \textit{If it matches, join the current cluster.}}
		\STATE  $\hat{x} \gets x_i$ , $\hat{y} \gets y_i$, $\mu_{\text{joint}} \gets \mu_{\text{joint}} + \mu_{x_i}^{j_{\max}}$, $n \gets n + 1$ 
		\ELSE
		\STATE \textcolor{gray}{$ // $ \textit{If it doesn’t match, start the next cluster.}}
		\STATE $ C_k^{\text{num}} \gets n$, $ C_k^{\text{ave}} \gets \frac{\mu_{\text{joint}}}{n} $, $\mu_{\text{joint}} \gets \mu_{x_i}^{j_{\max}}$, $\hat{x} \gets x_i$ , $\hat{y} \gets y_i$, $ k \gets k + 1 $, $ n \gets 1$
		\ENDIF
		\ENDFOR
		\STATE $ k\gets k + 1 $, $ C_k^{\text{num}}\gets n$, $ C_k^{\text{ave}}\gets \frac{\mu_{\text{joint}}}{n} $, $ k_{\max} \gets \mathop{\arg\max}_i (C_i^{\text{ave}})$
        \STATE$ s \gets \sum_{t=1}^{k_{\max}}{C_t^{\text{num}}}$, $ e \gets \sum_{t=1}^{k_{\max}}{C_t^{\text{num}}} + C_{k_{\max}}^{\text{num}} $
		\RETURN $\mathcal{I}=\{x_s,x_s+1,\dots,x_e\}$
	\end{algorithmic}
	\label{alg1}
\end{algorithm}
The clustering is constrained by three rules: (1) the rows of $V_{\text{cand}}$s within a cluster must be adjacent; (2) the distance between their $ y_t $ values must be less than $ T_c $; (3) the distance between their $ y_t + \mu_{x_t}^{j_{\max}} $ values must also be less than $ T_c $. {\color{black} By applying the above procedure, multiple clusters can be obtained. The length of each cluster is subsequently computed, and the cluster with the maximum length is selected as the final $S_{\core}$. 
The set $\mathcal{I} = \{ \mathcal{I}_i \}_{i=1}^{n_{\core}}, \mathcal{I}_i \in \{1, \dots, M\}$ records the indices of rows belonging to $S_{\core}$, from which the starting coordinates $\mathcal{P}$ and lengths $\mathcal{L}$ are directly extracted from $C_{\text{start}}$ and $C_{\text{len}}$. The complete procedure is depicted in Algorithm~\ref{alg1}.}

The robustness of $S_{\core}$ is attributed to two factors. \textbf{1)} We exclusively focus on $S_{\core}$, ensuring that unconnected noisy pixels exert no influence on its extraction. \textbf{2)} The use of RLC guarantees that only consecutive runs of identical pixel values are considered valid candidates $V_{\text{cand}}$s. Since noise typically manifests as isolated pixels, such disturbances are inherently eliminated during the construction of $V_{\text{cand}}$. 

\begin{figure}[t]
\vspace{-0.9em}
\centering
\subfloat[]{
\centering
\includegraphics[width=0.23\textwidth]{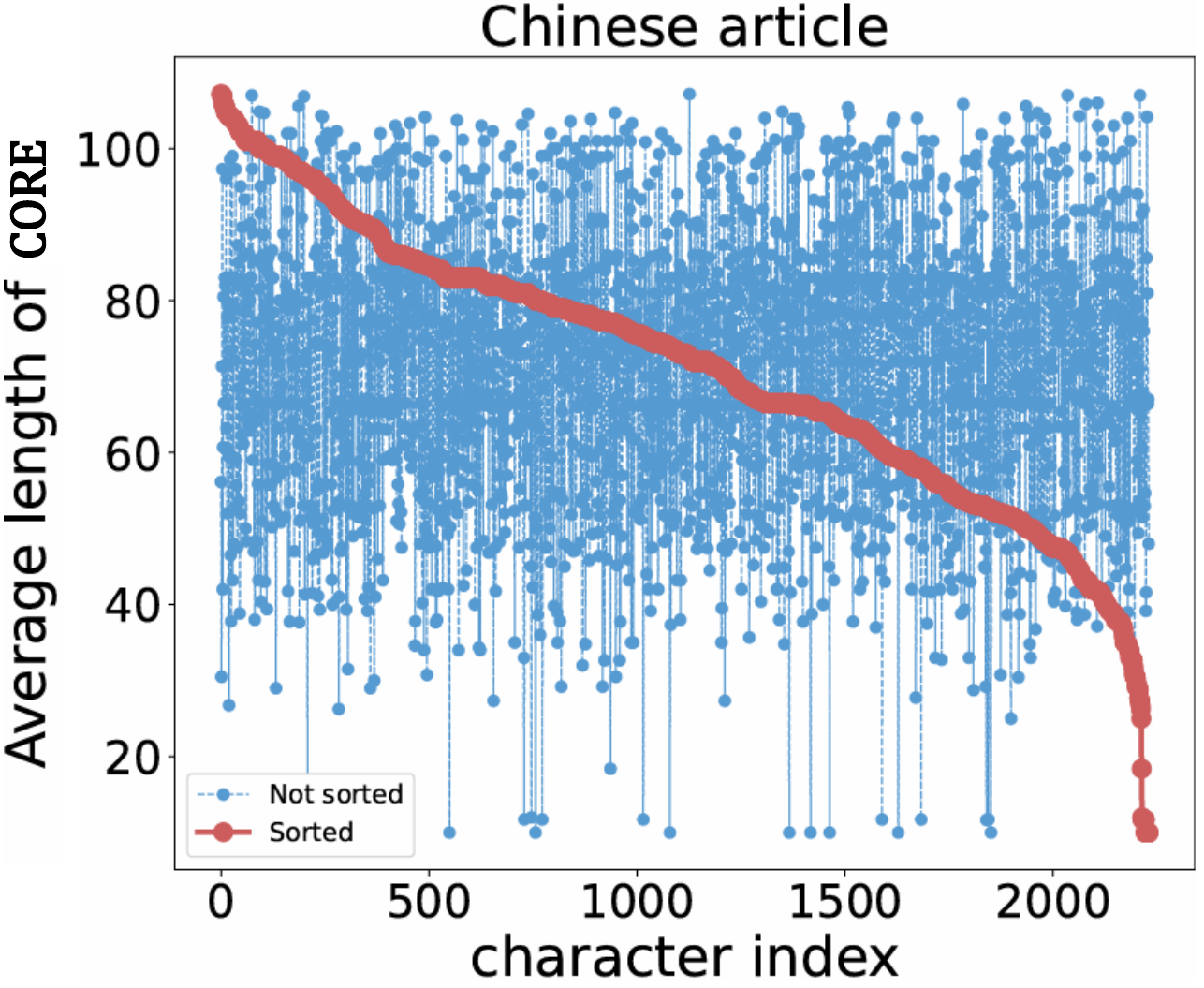}
}
\subfloat[]{
\centering
\includegraphics[width=0.23\textwidth]{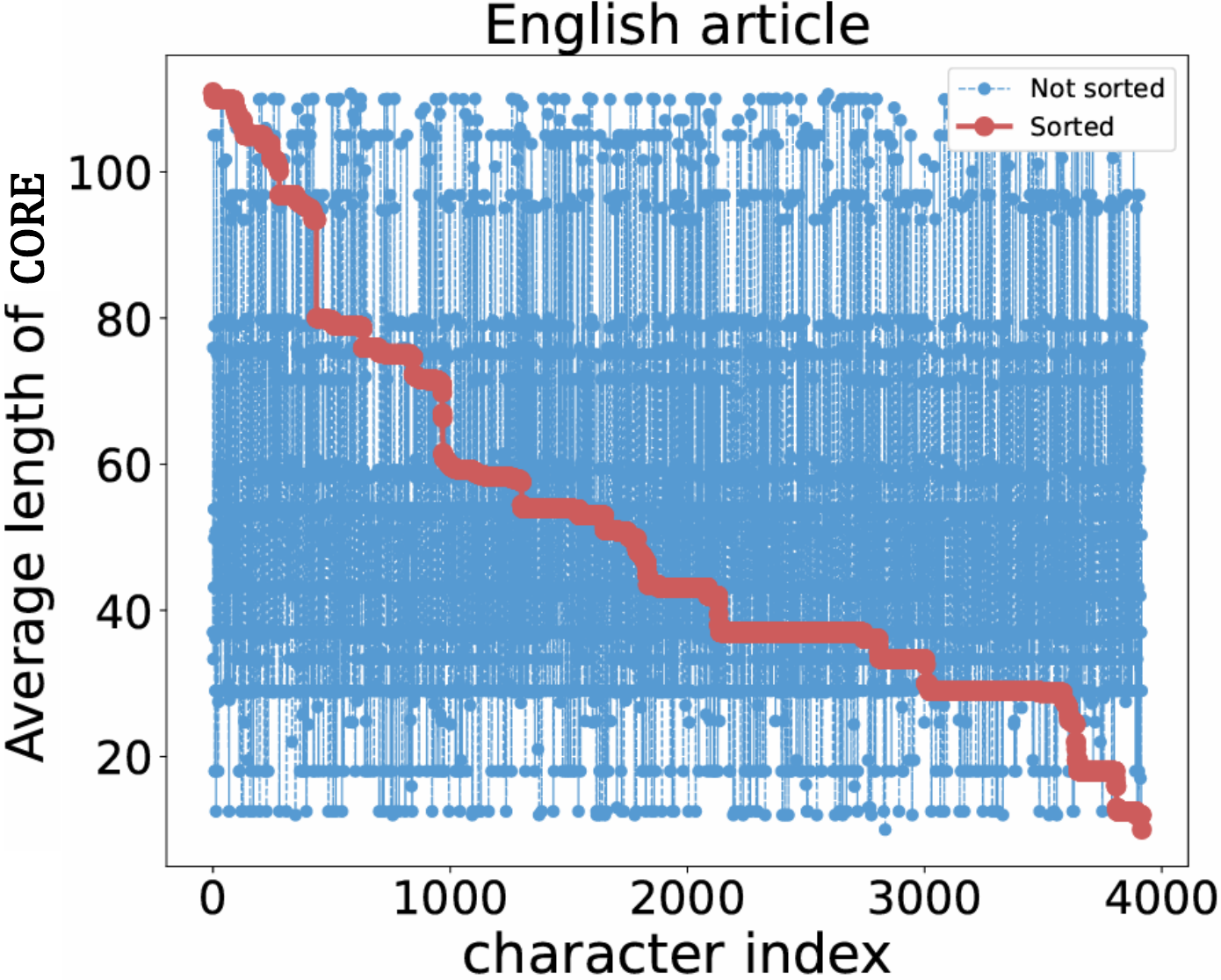}
}
\caption{Analysis of the distribution of lengths of $ S_{\core} $s in Chinese and English documents: (a) Average lengths of $ S_{\core} $s in Chinese document (b) Average lengths of $ S_{\core} $s in English document. }
\label{fig_6}
\vspace{-0.5em}
\end{figure}

\vspace{0.3em}
\noindent \textbf{Component 2: Character Selection.}
As is known, characters from different languages, fonts, and punctuation systems across the world possess unique pixel distributions. In particular, the distributions of their \textsc{core}s also differ significantly. As shown in Figure \ref{fig_6}, the blue points represent the unsorted average lengths of $ S_{\core} $s, while the red points show these average lengths after sorting, revealing a distribution ranging from long to short.
Longer lengths correlate with greater robustness. Consequently, certain characters prove unsuitable for information embedding due to insufficient robustness.  For example, the $ S_{\core} $ extracted from character `,' is too short to withstand distortions during transmission.
To address this limitation, we propose a character selection strategy. 

{\color{black} Recall that $\mathcal{L} = \{ \ell_i \}_{i=1}^{n_{\core}}$ recordes the length of each sequence in $S_{\core}$. The overall length of $S_{\core}$, denoted as $\ell^{\core}$, is the average length across all sequences and is defined below: 
\begin{equation}
    \ell^{\core}= \frac{\sum_{i=1}^{n_\core}\ell_i}{n_{\core}},
    \label{ls}
\end{equation}
Let $ C = \{c_i\}_{i=1}^{n_c}$ be the set of all characters in a text image, where $n_c$ is the total number of characters contained in the image. The \textsc{core} lengths of all characters are stored in the set $L_{\core} = \{ \ell^{\core}_i\}_{i=1}^{n_c}$.} 
We establish a threshold $ T_{\lambda} $ to determine character eligibility. Notably, $ T_{\lambda} $ is not fixed but calculated based on the lengths of $ S_{\core} $s in the document. 
To calculate $ T_{\lambda} $, we first sort all lengths of $ S_{\core} $s from shortest to longest according to $ L_{\core} $,  with the sorted values represented by $\tilde{L}_{\core} = \{\tilde{\ell}^{\core}_i\}_{i=1}^{n_c}$. 
{\color{black}Assuming $ \lambda \in (0, 1]$ is a user-chosen percentile, we calculate $ T_{\lambda} $ as follows:
\begin{equation}
k_\lambda = \lceil n_c \cdot \lambda \rceil, \quad T_{\lambda} = \tilde{\ell}^{\core}_{k_\lambda}. 
\label{tlamda}
\end{equation}
We set $\lambda=0.2$ in this paper.}

Let W = $ \{w_i\}_{i = 1}^{L_w}$ be a watermark sequence with length $ L_w $ to be embedded. We first assess whether the length of $ S_{\core} $ for the current character $ c_i $ exceeds $ T_{\lambda} $; If it does, we embed one bit of information in $ c_i $; if not, we apply the same evaluation to the next character. This process continues until all information has been embedded. 

\vspace{0.3em}
\noindent \textbf{Component 3: Embedding Mechanism.}
\label{watermark embed}
In our proposed scheme, the embedding feature for watermarking is defined as the size of $S_{\core}$, \ie, $n_{\core}$. To embed a bit “0”, we ensure $n_{\core}$ exceeds a predetermined threshold $ T_{\Delta} $ by $\beta$. Conversely, to embed a bit "1", we constrain $n_{\core}$ to be less than $ T_{\Delta} $ by $\beta$. Here, $\beta$ represents the watermark embedding strength. 
{\color{black} Let $\tilde{n}_{\core}$ denote the thickness of $ S_{\core} $ after embedding. The watermark embedding process for each character can then be expressed mathematically as:
\begin{equation}
	{\tilde{n}_{\core}} = \begin{cases}
		T_{\Delta} - \beta, &\text{if $T_{\Delta} - n_{\core} \le \beta$ and $ w_i = 1$}\\
		n_{\core}, &\text{if $T_{\Delta} - n_{\core} > \beta$ and $ w_i = 1$}\\
		T_{\Delta} + \beta + 1, &\text{if $T_{\Delta} - n_{\core} \ge \beta$ and $ w_i = 0$}\\
		n_{\core}, &\text{if $T_{\Delta} - n_{\core} < \beta$ and $ w_i = 0$}\\
	\end{cases}.
	\label{eq8}
\end{equation}
}

Specifically, we designate a specific text line as the baseline for calculating the threshold $T_{\Delta}$, defined as the mean of $n_{\core}$ values over all characters in this line. Characters within this baseline are not subject to watermark embedding.
We satisfy the condition specified in Eq.~\ref{eq8} by either reducing or expanding $ S_{\core} $. In particular, human perceptual factors are considered by analyzing neighboring pixels of $S_{\core}$ to ensure that modifications do not introduce visually perceptible differences. The methods for reducing or expanding are discussed separately below.

\subsubsection{Reducing $S_{\core}$}
To ensure that $ C_{\max}^{\text{ave}} $ remains the largest value among $ C_{\text{ave}}$ after reducing, we select the shorter candidate vectors on both sides of $ S_{\core} $ and convert their black pixels to white. Using the horizontal direction as an example, the reducing process follows these steps:{\color{black}
\begin{enumerate}[\hspace{2.95em}]
	\item[Step 1:] Compare $ \ell_1 $ and $ \ell_{n_{\core}} $ to identify the shorter side. Let $k$ denote the index of the shorter segment, i.e., 
    $k = 1$ if $\ell_1$ is shorter and $k = n_{\core}$ if $\ell_{n_{\core}}$ is shorter. 
	\item[Step 2:]
    Use row $\mathcal{I}_{k+d}$ as the reference, where $d = -1$ if $k=1$ and $d=+1$ if $k=n_{\core}$. 
    A black pixel in row $\mathcal{I}_k$ is deemed `flippable' if the adjacent pixel in the reference row $\mathcal{I}_{k+d}$ is white. 
    After flipping, set $k \leftarrow k+1$ if $k=1$; set $k \leftarrow k-1$ if $k=n_{\core}$. 
	\item[Step 3:] Repeat the operation in Step 2 for $|T_{\Delta} - n_{\core}|+1$ iterations to complete the reducing process.
\end{enumerate}
}

\subsubsection{Expanding $S_{\core}$} In contrast to reducing, pixels outside $S_{\core}$ are converted to black in order to expand $S_{\core}$. The process of expanding is listed as follows:{\color{black}
\begin{enumerate}[\hspace{2.95em}]
	\item[Step 1:] Determine the longer side by comparing $ \ell_1 $ and $ \ell_{n_{\core}} $.
	\item[Step 2:]If $\ell_1$ is longer, flip all white pixels to black within the region bounded by x-coordinates ranging from $ \mathcal{I}_1 - |T_{\Delta} - n_{\core}| - 1$ to $ \mathcal{I}_1 - 1 $ and y-coordinates ranging from $y_{\mathcal{I}_1}$ to $y_{\mathcal{I}_1} + \ell_1 $.
    Alternatively, if $ \ell_{n_{\core}} $ is greater, flip all white pixels to black within the region bounded by x-coordinates ranging from $ \mathcal{I}_{n_\core} + 1$ to $ \mathcal{I}_{n_\core} + |T_{\Delta} - n_{\core}| + 1 $ and y-coordinates ranging from $y_{\mathcal{I}_{n_\core}}$ to $y_{\mathcal{I}_{n_\core}} + \ell_{\mathcal{I}_{n_\core}}$.
\end{enumerate}
}

\vspace{0.3em}
\noindent \textbf{Component 4: Independent Embedding Strength Modulator.}
\label{ES modulator}
\begin{figure}[t]
\includegraphics[width=0.49\textwidth]{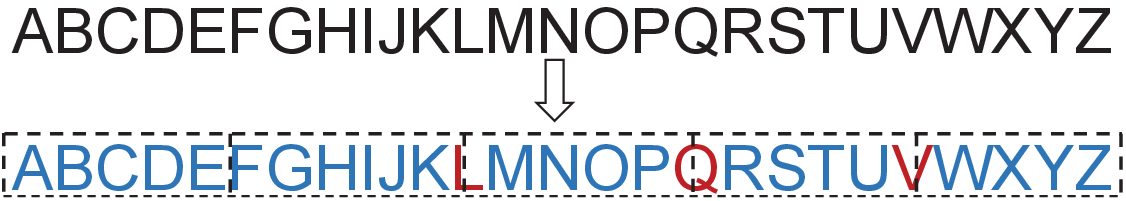}
\centering
\caption{The model of character segmentation. “Complete characters” in blue: A, B, C, D, E, F, G, H, I, J, K, M, N, O, P, R, S, T, U, W, X, Y, Z. “Incomplete characters” in red: L, Q, V.}
\label{fig_7}
\vspace{-0.5em}
\end{figure}
To enhance robustness when working with small font sizes, we propose a plug-and-play embedding strength modulator (ES modulator). 
First, we extract all text lines from the text image and segment each line into $ N_s $ sub-lines of equal length. Figure \ref{fig_7} illustrates our independent embedding strength modulator, where $N_s = 5$. Characters that remain intact during segmentation are termed "complete characters" (shown in blue), while those that are divided are termed "incomplete characters" (shown in red). Let sub-lines be denoted as $ B_i, i=1,...,N_s $, we embed the same bit $ w_i $ in all “complete characters” within each $ B_i $.  This approach allows the ES modulator to adaptively increase the number of "complete characters" to enhance robustness as font size decreases. Furthermore, we argue that ES modulator is highly modular and method-independent and it can be applied to various methods to improve robustness, which will be demonstrated in Section~\ref{experiment}. 

\subsection{Watermark Extraction}
To extract the hidden message from a text image, we first segment all characters and extract $ S_{\core} $s of each one. Next, robust characters are selected as described in Section \ref{watermark embed}. In the extraction stage, let $\hat{T}_\Delta$ denote the extraction threshold and $\hat{n}_{\core}$ denote the size of $S_{\core}$. The watermark bit can be extracted by comparing $\hat{n}_{\core}$ with $\hat{T}_{\Delta}$. The watermark detection process can be formulated as follows:
\begin{equation}
w_i = \begin{cases}
1, & \text{if $ \hat{T}_\Delta - \hat{n}_{\core}\ge 0$}\\
0, & \text{if $ \hat{T}_\Delta - \hat{n}_{\core} < 0$}\\
\end{cases}.
\end{equation}

\begin{figure*}
\centering
\includegraphics[width=0.95\textwidth]{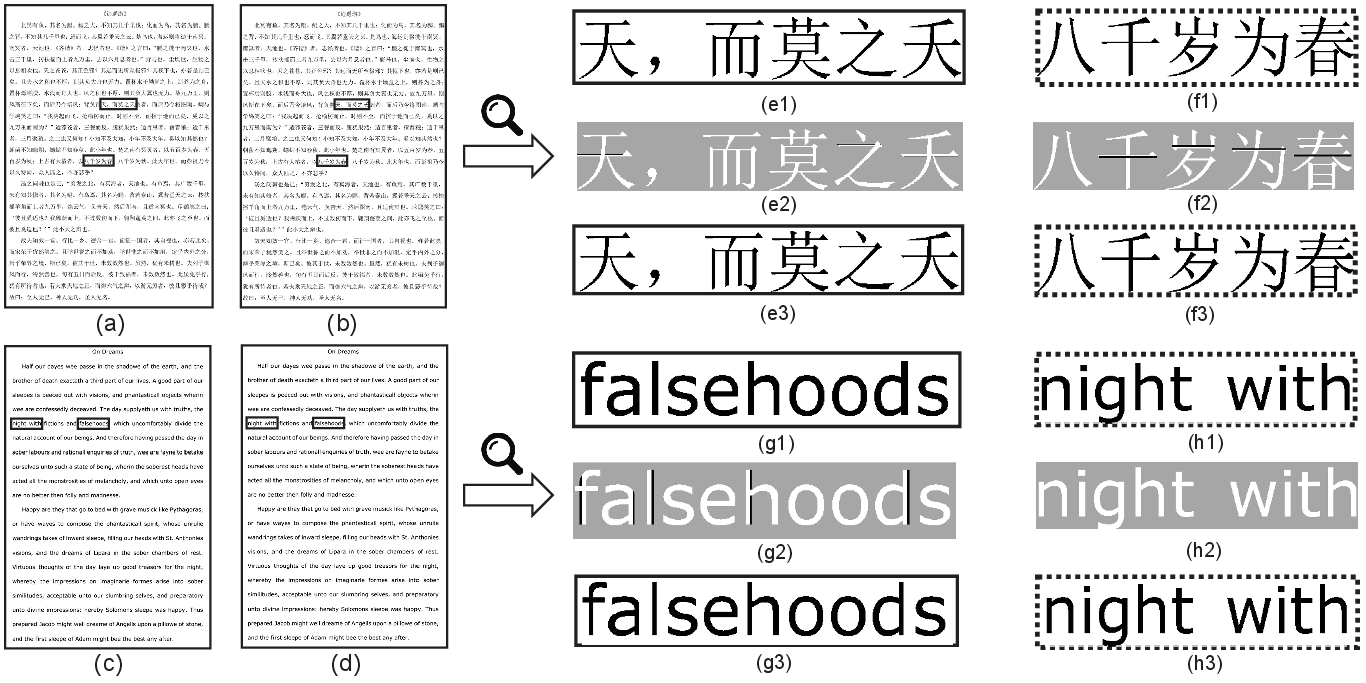}
\caption{Data hiding in English text document image and Chinese text document image: (a) original copy (Chinese); (b) marked copy (Chinese); (c) original copy (English); (d) marked copy (English); (e1, f1) magnified partial of (a); (e2) difference between (e1) and (e3) (shown in black); (e3) magnified partial of (b) (after reducing $S_{\core}$); (f2) difference between (f1) and (f3) (shown in black); (f3) magnified partial of (b) (after expanding $S_{\core}$); (g1, h1) magnified partial of (c); (g2) difference between (g1) and (g3) (shown in black); (g3) magnified partial of (c) (after reducing $S_{\core}$); (h2) difference between (h1) and (h3) (shown in black); (h3) magnified partial of (d) (after expanding $S_{\core}$);}
\label{fig_9}
\end{figure*}

\begin{figure*}[t]
\centering
\includegraphics[width=0.95\textwidth]{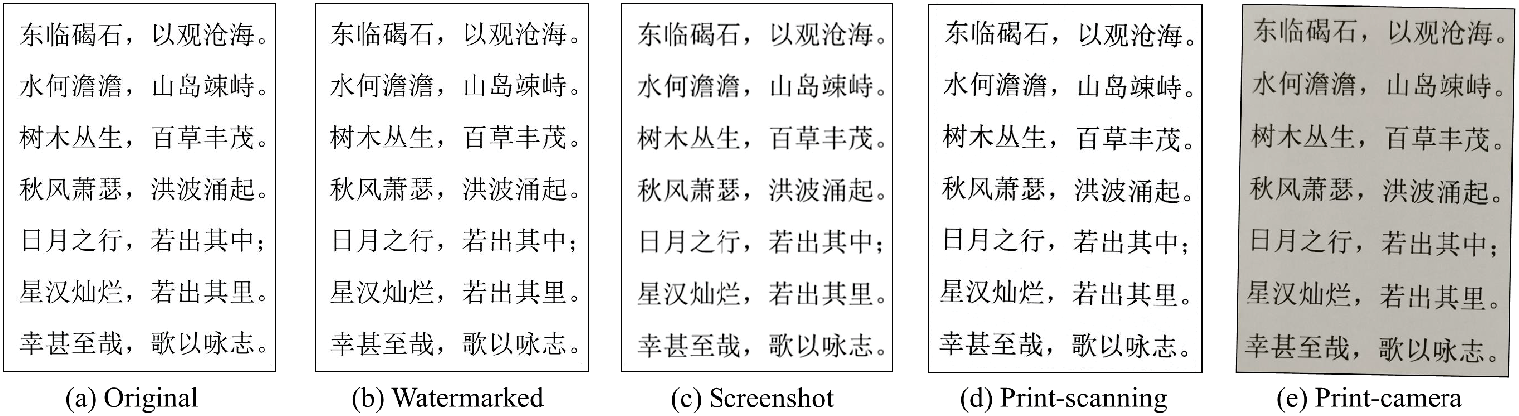}
\caption{Display of the original text image, text image with watermark, and effects after different distortions (eg, SongTi, 14pt).}
\label{display}
\vspace{-0.5em}
\end{figure*}

\section{Experimental Results}
\label{experiment}
\subsection{Settings}

\noindent
\textbf{Datasets.}
We evaluate our method on both English and Chinese fonts. The English fonts include Calibri, Arial, Times New Roman (TNR), Segoe UI, and Verdana, while the Chinese fonts comprise SimSun, SimHei, DengXian, and Microsoft YaHei (Msyh). For Calibri and SimSun, we test eight font sizes (10-24pt in 2pt increments). We further assess CoreMark's generalizability across six foreign language fonts (Arabic, French, Japanese, Korean, Russian, Spanish) to demonstrate cross-linguistic robustness.

\vspace{0.3em}
\noindent \textbf{Baseline Selection.}
We compare DiffMark with two representative image-based text watermarking methods (Wu04 \cite{wu2004data} and Yang23 \cite{yang2023language}) and include results from the state-of-the-art (SOTA) font-based method for reference. All methods undergo identical experimental conditions including equipment and environmental settings to ensure fair evaluation.

\vspace{0.3em}
\noindent \textbf{Settings for Text Watermarking Methods.}
For CoreMark, we set $ N_s =9$ and $ T_c =10$.
For Wu04 \cite{wu2004data}, we set $ B =200$ and $ Q=500 $, while for Yang23 \cite{yang2023language}, we set $r = 12$ and $N = 2$. As shown in Figure~\ref{revisitBalance}, we intentionally select larger step sizes for baseline methods to maximize their robustness, thereby providing more rigorous comparison that highlights CoreMark's superior performance.
The parameters settings for AutoStegaFont are the same as those in \cite{yang2023autostegafont}.  

\vspace{0.3em}
\noindent \textbf{Evaluation Metrics.}
For robustness, we employ average extraction accuracy ($ ACC $) defined in Eq.~\ref{ACC}, where higher values indicate greater robustness. All $ ACC $ values are reported without error correction encoding. For invisibility assessment, we adopt PSNR and SSIM as evaluation metrics. Larger values of PSNR and SSIM indicate better imperceptibility. These metrics are measured between original and watermarked binary images rather than grayscale or color images. 

\subsection{Main Results}

\noindent \textbf{Watermark Imperceptibility.}
\label{imperceptibility}
Figure~\ref{fig_9}(a) and Figure~\ref{fig_9}(c) show the original Chinese and English documents, respectively.
Figure~\ref{fig_9}(b) and Figure~\ref{fig_9}(d) are the corresponding watermarked documents. On the right side of Figure~\ref{fig_9}, we enlarge portions of the original images and the marked images and represent their differences with black pixels. We can see that some characters are not modified, such as the second character “,” in Figure~\ref{fig_9}(e2) and the first character “n” in Figure~\ref{fig_9}(h2). The reasons are: 1) The character was not selected for embedding, as explained in Section~\ref{watermark embed}; 2) $ \Delta > 0 $ when embedding “1”, and $ \Delta < 0 $ when embedding “0”, as described in Eq.~\ref{eq8}, respectively. 
Table~\ref{different_font_psnr} reports a quantitative comparison of visual quality between original and watermarked images across a variety of font styles. The proposed CoreMark consistently outperforms three baselines across both English and Chinese font families. Specifically, CoreMark achieves the highest PSNR and SSIM scores in all cases, with PSNR gains of up to 11.31 dB and SSIM values exceeding 0.999 in most cases, indicating minimal visual distortion. These results highlight the superior imperceptibility of CoreMark across a wide range of font styles and demonstrate its effectiveness in preserving visual quality.

\begin{table*}[t]
\centering
\renewcommand{\arraystretch}{1.1}
\caption{Robustness ($ ACC(\%) $) against screenshots of text with different sizes in different languages.}
\vspace{-0.3em}
\label{different_font_psnr}%
\setlength{\tabcolsep}{3mm}
\begin{center}
    \begin{tabular}{c|c|ccccc|cccc}
    \hline
    \multirow{2}{*}{\raisebox{-0.6ex}[0pt][0pt]{Metrics}} & \multirow{2}{*}{\raisebox{-0.6ex}[0pt][0pt]{Method}} & \multicolumn{5}{c|}{English}&\multicolumn{4}{c}{Chinese}\\
    & & Calibri & Arial & TNR & Segoe UI & Verdana & Simsun & Simhei & DengXian & Msyh \\
    \hline
    \multirow{4}{*}{\raisebox{-0.6ex}[0pt][0pt]{PSNR}} 
    & Wu04 \cite{wu2004data} & 28.50 & 28.64& 28.84& 29.94& 28.81& 28.70& 28.72&28.63&28.25 \\
    & T\_BB \cite{tan2019print} & 29.13& 28.20&29.56 & 29.40& 27.86& 28.77&27.35 & 28.45&26.98\\
    &Yang23\cite{yang2023language} &26.31 &27.41&29.08 &28.59& 26.71 & 24.04 & 25.86 &26.45 &25.60\\
    & CoreMark (Ours) &\textbf{37.62} & \textbf{38.31} & \textbf{39.19} & \textbf{39.50} & \textbf{37.07} & \textbf{34.57}& \textbf{36.51} & \textbf{36.38} & \textbf{35.12}\\
    \hline
    \multirow{4}{*}{\raisebox{-0.6ex}[0pt][0pt]{SSIM}} 
    & Wu04 \cite{wu2004data} & 0.9931& 0.9930& 0.9936& 0.9946& 0.9934& 0.9936& 0.9928& 0.9932&0.9918\\
    & T\_BB \cite{tan2019print} & 0.9958&0.9950& 0.9960& 0.9964&0.9948& 0.9947& 0.9928& 0.9942&0.9923 \\
    &Yang23\cite{yang2023language} & 0.9901 & 0.9880& 0.9898 & 0.9870 & 0.9916 &0.9828& 0.9936 & 0.9945 & 0.9912 \\
    & CoreMark (Ours) &\textbf{0.9991}& \textbf{0.9990} & \textbf{0.9990} & \textbf{0.9982} & \textbf{0.9991}& \textbf{0.9988}& \textbf{0.9999} & \textbf{0.9995} & \textbf{0.9988}\\
    \hline
\end{tabular}
\end{center}
\vspace{-0.3em}
\end{table*}

\begin{table*}[t]
\centering
\renewcommand{\arraystretch}{1.1}
\caption{Robustness ($ ACC(\%) $) against screenshots of text with different sizes in different languages.}
\vspace{-0.3em}
\label{different-fontsize}%
\setlength{\tabcolsep}{3.5mm}
\begin{center}
\begin{tabular}{c|c|c|cccccccc}
\hline
\multirow{2}{*}{\raisebox{-0.6ex}[0pt][0pt]{Language}} &\multirow{2}{*}{\raisebox{-0.6ex}[0pt][0pt]{Attack}}& \multirow{2}{*}{\raisebox{-0.6ex}[0pt][0pt]{Method}} & \multicolumn{8}{c}{fontsize}\\
& & & 10pt & 12pt & 14pt & 16pt & 18pt & 20pt & 22pt & 24pt \\
\cline{1-11}
\multirow{12}{*}{\raisebox{-0.6ex}[0pt][0pt]{Chinese}} &\multirow{4}{*}{\raisebox{-0.6ex}[0pt][0pt]{Screenshots}}
    & Wu04 \cite{wu2004data} &48.63 & 50.12 & 46.79 & 52.45 & 49.86 & 53.30 & 47.58 & 51.03   \\
&&Yang23\cite{yang2023language} & 78.61 & 87.02 & 94.51 & 93.26 & 95.43 & 94.99 & 94.41 & 95.28 \\
&& AutoStegaFont \cite{yang2023autostegafont} &60.76& 62.72&64.77& 67.06&67.53& 67.81& 70.98&72.73 \\
&& CoreMark (Ours) & \textbf{94.59} & \textbf{96.99} & \textbf{97.30} & \textbf{97.35} & \textbf{97.38} & \textbf{97.59} &\textbf{97.67} & \textbf{97.72} \\
\cline{2-11}
&\multirow{4}{*}{\raisebox{-0.6ex}[0pt][0pt]{Print-scan}}
& Wu04 \cite{wu2004data}& 50.95 & 48.14 & 53.76 & 47.90 & 49.22 & 51.61 & 45.98 & 52.07\\
&&Yang23\cite{yang2023language} & 72.01 & 85.40 & 90.80 & 90.52 & 95.53 & 95.80 & 95.82 & 94.49 \\
&& AutoStegaFont \cite{yang2023autostegafont} &69.55&70.85&70.96&71.07&71.07&71.59&72.07&72.51\\
&& CoreMark (Ours) & \textbf{93.38} & \textbf{93.59} & \textbf{94.00} & \textbf{95.38} & \textbf{95.59} & \textbf{96.19} & \textbf{96.38} & \textbf{96.80}\\
\cline{2-11}
&\multirow{4}{*}{\raisebox{-0.6ex}[0pt][0pt]{Print-camera}}
& Wu04 \cite{wu2004data}& 49.05 & 48.91 & 52.61 & 47.37 & 50.77 & 53.14 & 46.88 & 51.29 \\
&&Yang23\cite{yang2023language} & 71.55 & 83.24 & 89.05 & 90.23 & 94.01 & 94.44 & 95.18 & 95.37\\
&& AutoStegaFont \cite{yang2023autostegafont} & 55.97& 57.23& 57.56& 57.63&60.73& 64.97&66.09& 67.26\\
&& CoreMark (Ours) & \textbf{91.82} & \textbf{92.88} & \textbf{94.19} & \textbf{94.89} & \textbf{95.17} & \textbf{96.35} & \textbf{96.68} & \textbf{96.90} \\
\hline
\multirow{12}{*}{\raisebox{-0.6ex}[0pt][0pt]{English}} &\multirow{4}{*}{\raisebox{-0.6ex}[0pt][0pt]{Screenshots}}
& Wu04 \cite{wu2004data} & 48.92& 51.36& 46.25& 50.78& 54.31& 49.87& 47.59&52.44 \\
&&Yang23\cite{yang2023language} & 81.37 & 83.44 & 84.36 & 84.58 & 84.67 & 84.73 & 85.43 & 85.94  \\
&& AutoStegaFont \cite{yang2023autostegafont} & 55.12 & 58.97 & 59.92 & 66.55 & 68.97 & 70.58 & 73.02 & 77.41\\
&& CoreMark (Ours) & \textbf{95.25} & \textbf{97.26} & \textbf{97.50} & \textbf{98.25} & \textbf{98.78} & \textbf{99.75} & \textbf{100} & \textbf{100} \\
\cline{2-11}	
&\multirow{4}{*}{\raisebox{-0.6ex}[0pt][0pt]{Print-scan}} 
& Wu04 \cite{wu2004data} &48.72 & 51.39 & 49.85 & 52.04 & 46.91 & 50.27 & 47.68 & 53.11 \\
&&Yang23\cite{yang2023language} & 80.51 & 82.33 & 83.91 & 84.29 & 84.55& 84.70 & 84.92 &85.81 \\
&& AutoStegaFont \cite{yang2023autostegafont} &51.85 & 50.01 & 53.85 & 52.89 & 56.74 & 54.24 & 48.08 & 58.66 \\
&& CoreMark (Ours) & \textbf{95.29} & \textbf{95.79} & \textbf{95.89} & \textbf{96.18} & \textbf{97.64} & \textbf{97.79} & \textbf{97.99} & \textbf{98.38}  \\
\cline{2-11}
&\multirow{4}{*}{\raisebox{-0.6ex}[0pt][0pt]{Print-camera}} 
& Wu04 \cite{wu2004data}& 49.34 & 52.88 & 47.21 & 50.67 & 46.85 & 53.42 & 48.96 & 51.10 \\
&&Yang23\cite{yang2023language} & 79.01 & 82.33& 83.35& 83.94& 84.48 & 84.53 & 84.68 & 85.29\\
&& AutoStegaFont \cite{yang2023autostegafont} & 53.91 & 51.49 & 57.74 & 52.31 & 55.80& 54.85 & 53.91&53.99\\
&& CoreMark (Ours) & \textbf{93.39} & \textbf{94.59} & \textbf{95.11} & \textbf{95.28} & \textbf{96.36} & \textbf{97.19} & \textbf{97.60} & \textbf{97.79}\\
\hline
\end{tabular}
\end{center}
\vspace{-0.3em}
\end{table*}

\vspace{0.3em}
\noindent \textbf{Watermark Robustness (against Screenshots).}
\label{sectionscreen}
Text images are often propagated via screenshots, with distortions primarily arising from the down-sampling of the screen. As shown in Table~\ref{different-fontsize}, no matter Chinese fonts or English fonts, all $ ACC $s of our CoreMark exceed 94\%, which significantly outperform the baseline methods. Even in the case of 10pt, the extraction success rates of CoreMark for Chinese and English fonts are still above 94\% and 95\%, respectively, which indicates that CoreMark can achieve satisfactory robustness even with smaller fontsizes. In contrast, the performance of baseline methods exhibits noticeable limitations across both Chinese and English texts.
Wu04 and AutoStegaFont yield relatively low extraction accuracy, with their lack of robustness analyzed in detail in Sections~\ref{strategy1} and \ref{strategy3}, respectively. Although Yang23 demonstrates improved resilience against screenshots, achieving accuracies ranging from 78.61\% to 95.43\% for Chinese and 81.37\% to 93.83\% for English, its performance remains unstable, particularly under smaller font sizes.

\begin{table}
\caption{Comparison results of the robustness of CoreMark without and with IES modulator in the printing and scanning process under different device combinations.}
\vspace{-0.3em}
\label{equipment}
\centering
\renewcommand{\arraystretch}{1.1}
\setlength{\tabcolsep}{1mm}
\begin{tabular}[t]{c|c|ccc}
\hline
\multirow{2}{*}{\raisebox{-0.6ex}[0pt][0pt]{}} & \multirow{2}{*}{\raisebox{-0.6ex}[0pt][0pt]{Printer}} & \multicolumn{3}{c}{Scanner}\\
&& Epson V30 & M132nw & Canon LiDE400 \\
\hline
\multirow{2}{*}{\raisebox{-0.6ex}[0pt][0pt]{CoreMark w/o. IES}}& M132nw & 96.69 & 94.29 &  95.80\\
& M17w & 95.65 & 93.91 & 95.13 \\
\hline
\multirow{2}{*}{\raisebox{-0.6ex}[0pt][0pt]{CoreMark w/. IES}} & M132nw & 98.99 & 96.56 & 97.39 \\
&M17w & 97.44 & 96.02 & 97.01 \\
\hline
\end{tabular}
\vspace{-0.3em}
\end{table}

\begin{table*}[t]
\renewcommand{\arraystretch}{1.1}
\caption{Qualitative Analysis of CoreMark in Different Fonts.}
\vspace{-0.3em}
\label{differentfont-robustness}
\centering
\setlength{\tabcolsep}{3.5mm}
\begin{tabular}{c|c|ccc|cccc}
\hline
\multirow{2}{*}{\raisebox{-0.6ex}[0pt][0pt]{ }}&\multirow{2}{*}{\raisebox{-0.6ex}[0pt][0pt]{Method}} & \multicolumn{3}{c|}{Chinese 14pt} & \multicolumn{4}{c}{English 18pt}\\
& &Heiti & Dengxian & Msyh & Arial &TNR & Segoe UI & Verdana\\
\hline
\multirow{3}{*}{\raisebox{-0.6ex}[0pt][0pt]{\makecell[c]{$ACC  $\\(Screenshot)}}} 
& Wu04 \cite{wu2004data}&51.28&51.46&53.39&52.16&51.89&50.64&50.35\\
&Yang23\cite{yang2023language} & 66.25 & 76.79 & 60.12 & 82.01 & 96.02 & 85.18 & 84.43\\ 
& CoreMark (Ours)& \textbf{96.71} & \textbf{98.19} & \textbf{97.12} & \textbf{99.29} & \textbf{99.39} & \textbf{98.67} & \textbf{98.92}\\ 
\hline
\multirow{3}{*}{\raisebox{-0.6ex}[0pt][0pt]{\makecell[c]{$ACC  $\\(Print-scan)}}} 
& Wu04 \cite{wu2004data}&49.73&52.18&46.57&50.89&48.34&54.66&45.92\\
&Yang23\cite{yang2023language} & 69.26 & 78.62& 66.47 & 80.41 & 96.24 & 79.11 & 86.69 \\
& CoreMark (Ours) & \textbf{91.09} & \textbf{94.55} & \textbf{95.49} & \textbf{97.83} & \textbf{97.47} & \textbf{96.81} & \textbf{97.25} \\ 
\hline
\multirow{3}{*}{\raisebox{-0.6ex}[0pt][0pt]{\makecell[c]{$ACC  $\\(Print-camera)}}} 
& Wu04 \cite{wu2004data}&47.28&51.67&49.85&50.12&53.49&46.93&48.77\\
&Yang23\cite{yang2023language}& 62.48 & 73.84 & 63.89 & 84.20& 94.39 & 77.67 & 80.12\\
& CoreMark (Ours) & \textbf{91.71} & \textbf{95.15} & \textbf{96.12}& \textbf{97.68}& \textbf{95.25}& \textbf{96.17}& \textbf{97.12}\\

\hline
\end{tabular}
\end{table*}
\begin{table*}[t]
\renewcommand{\arraystretch}{1.1}
\caption{Qualitative Analysis of CoreMark in Foreign Languages.}
\vspace{-0.3em}
\label{foreign}
\centering
\setlength{\tabcolsep}{4.5mm}
\begin{tabular}{c|c|cccccc}
\hline
&Method &Arabic & French & Japaneses & Korean &Russian & Spanish \\
\hline
\multirow{3}{*}{\raisebox{-0.6ex}[0pt][0pt]{\makecell[c]{$ACC  $\\(Screenshot)}}} 
& Wu04 \cite{wu2004data}&48.12&49.40&48.50&47.05&48.86&48.36\\
&Yang23\cite{yang2023language} & 59.63 & 79.58 & 82.83 & 77.67 & 92.06 & 85.24 \\
& CoreMark (Ours) & \textbf{90.75} & \textbf{96.49} & \textbf{95.84} & \textbf{97.13} & \textbf{97.99} & \textbf{98.36} \\  
\hline
\multirow{2}{*}{\raisebox{-0.6ex}[0pt][0pt]{\makecell[c]{$ACC  $\\(Print-scan)}}} 
& Wu04 \cite{wu2004data}&49.02&56.52&52.38&59.35&47.21&48.25\\
&Yang23\cite{yang2023language} & 52.22 & 65.56 & 63.05 & 70.01 & 68.33 & 79.11\\
& CoreMark (Ours) & \textbf{90.13} & \textbf{95.84}& \textbf{94.69} & \textbf{96.63} & \textbf{97.24} & \textbf{96.94}\\ 
\hline
\multirow{2}{*}{\raisebox{-0.6ex}[0pt][0pt]{\makecell[c]{$ACC  $\\(Print-camera)}}} 
& Wu04 \cite{wu2004data}&54.02&53.64&55.17&49.65&50.62&48.75\\
&Yang23\cite{yang2023language}& 55.37 & 71.90 & 69.44 & 73.01  & 73.55 & 76.58\\
& CoreMark (Ours) & \textbf{85.33} & \textbf{97.03} & \textbf{89.67} &\textbf{96.01}&\textbf{93.86} & \textbf{95.21}\\
\hline
\multirow{2}{*}{\raisebox{-0.6ex}[0pt][0pt]{PSNR (dB)}}
& Wu04 \cite{wu2004data}&30.58&28.50&28.32&29.76&28.18&28.35\\
&Yang23\cite{yang2023language} &23.38 & 30.72& 28.81 & 28.42 & 29.62 & 28.54 \\ 
& CoreMark (Ours) & \textbf{32.64} & \textbf{40.45} & \textbf{38.22} & \textbf{37.21} & \textbf{40.13} & \textbf{39.44} \\
\hline
\multirow{2}{*}{\raisebox{-0.6ex}[0pt][0pt]{SSIM}} 
& Wu04 \cite{wu2004data}&0.9955&0.9928&0.9925&0.9957&0.9919&0.9928\\
&Yang23\cite{yang2023language} & 0.9845 & 0.9966 & 0.9944& 0.9956& 0.9955& 0.9944 \\ 
& CoreMark (Ours) & \textbf{0.9978} & \textbf{0.9995} & \textbf{0.9990} & \textbf{0.9990} & \textbf{0.9994} & \textbf{0.9994}\\
\hline
\end{tabular}
\vspace{-0.3em}
\end{table*}

\vspace{0.3em}
\noindent \textbf{Watermark Robustness (against Print-scanning).}
\label{sectionscan}
Printers and scanners are widely used in daily document processing due to their accessibility and convenience. Printing generates physical document copies while scanning digitizes hardcopy materials for electronic storage, editing, or analysis. To verify the robustness against print-scan, we first print text images with different fontsizes on A4 paper using HP LaserJet Pro MFP M132nw at 600 dpi, then digitize them using Epson V30 SE scanner at 600 dpi. As shown in Table~\ref{differentfont-robustness}, our approach consistently achieves $>93$\% accuracy at 10pt for both Chinese and English texts, significantly outperforming baseline methods, particularly at smaller font sizes. To assess cross-device generalizability, we test CoreMark across multiple printer-scanner combinations using 50 randomly selected images with varied font sizes. Table~\ref{equipment} presents the extraction accuracy across two printer models at 600dpi (HP LaserJet Pro MFP M132nw and HP LaserJet Pro M17w) and three scanners at 600 dpi (Epson V30, M132nw, and Canon LiDE400). CoreMark shows strong robustness across all configurations, with further improvements when integrated with the ES modulator, confirming the effectiveness of the ES modulator in enhancing resilience against cross-device variations. 

\begin{table}[t]
\centering
\renewcommand{\arraystretch}{1.3}
\caption{Average Extraction Accuracy Under Varying Shooting Distances and Angles.}\label{shooting}%
\vspace{-0.3em}
\setlength\tabcolsep{1mm}{
\begin{tabular}{cccccccccc}
\hline
\multicolumn{5}{c}{Distances (cm)} & \multicolumn{5}{c}{Angles ($^\circ$)} \\
\cmidrule(lr){1-5}\cmidrule(lr){6-10}
10 &15&20&25&30&$5^\circ$ & $10^\circ$ & $15^\circ$ & $20^\circ$ & $25^\circ$\\
\hline
96.28&97.69&97.24&95.27&94.61&97.67&95.33&95.00&94.67&93.67\\
\hline
\end{tabular}
}
\vspace{-0.3em}
\end{table}

\vspace{0.3em}
\noindent \textbf{Watermark Robustness (against Print-camera).}
\label{sectioncamera}
Print-camera shooting is also a common method for capturing text content. We first print all the encoded text images on A4 paper at 600 dpi using the printer HP M132nw and take photos with a handheld cellphone HONOR 30S. The shooting distance is 20cm and the shooting angle is $0^\circ$. Subsequently, watermark extraction is performed on the images. As shown in Table~\ref{differentfont-robustness}, CoreMark maintains stable high accuracy, clearly outperforming the best competing baseline, Yang23 \cite{yang2023language}. When the Chinese font size exceeds 16pt, the $ACC$ of Yang23 surpasses 90\%. But when the font size is smaller, its extraction accuracy drops sharply.
To further assess the robustness of CoreMark against print-camera, we also evaluate it under other shooting conditions. The results, shown in Table~\ref{shooting}, demonstrate consistently high accuracy across all conditions, confirming the strong robustness of CoreMark against common real-world perturbations such as distance variation and angular deviation during image capture.

\begin{figure}[t]
\includegraphics[width=0.4\textwidth]{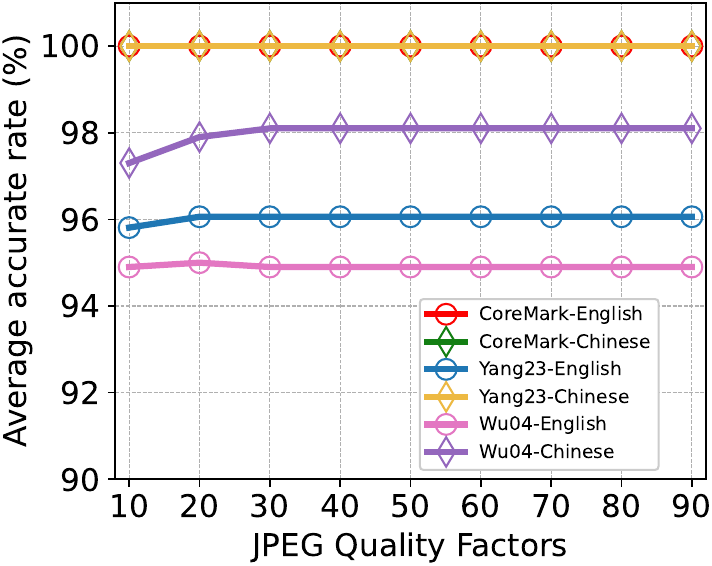}
\centering
\caption{Average extraction accuracy (\%) of three image-based methods (CoreMark, Yang23, and Wu04) under varying JPEG quality factors, evaluated on Calibri (English) and Simsun (Chinese) text images.}
\label{jpg}
\vspace{-0.3em}
\end{figure}

\begin{figure}[t]
\includegraphics[width=0.4\textwidth]{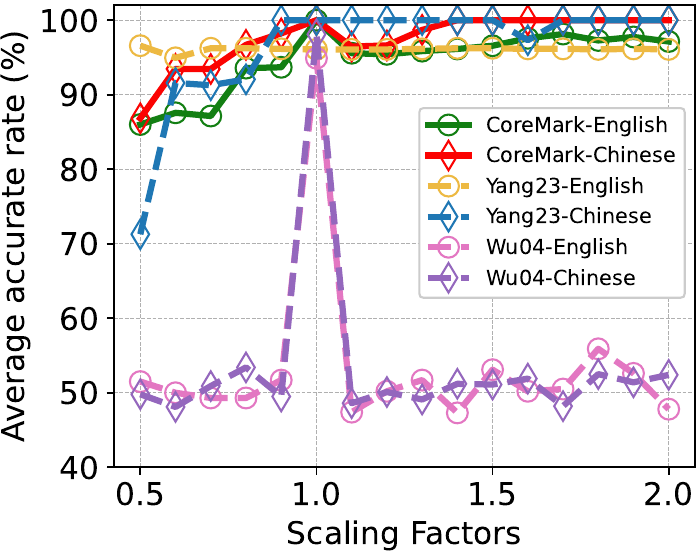}
\centering
\caption{Average extraction accuracy (\%) of three image-based methods (CoreMark, Yang23, and Wu04) under varying scaling factors, evaluated on Calibri (English) and Simsun (Chinese) text images. }
\label{scale}
\vspace{-0.3em}
\end{figure}

\vspace{0.3em}
\noindent \textbf{Watermark Robustness (against JPEG Compression and Scaling).}
JPEG compression and scaling are common operations in text image transmission. As shown in Figures~\ref{jpg} and \ref{scale}, we evaluate CoreMark against two baseline methods (Yang23 and Wu04), using English (Calibri) and Chinese (Simsun) text under varying compression qualities and scaling factors. For JPEG compression, CoreMark consistently achieves 100\% extraction accuracy across all compression levels in both English and Chinese. Yang23 also maintains stable performance, with average accuracy around 96\% for English and 100\% for Chinese. Notably, the extraction errors in Yang23 primarily result from the embedding instability discussed in Section~\ref{strategy2}, rather than from compression-induced degradation. Wu04 achieves an accuracy of approximately 98\% on Chinese text and around 95\% on English text. For Wu04, the observed extraction errors stem from embedding instability (as analyzed in Section~\ref{strategy2}) and the degradation effects introduced by JPEG compression.
For scaling transformations, CoreMark and Yang23 perform comparably well overall. However, Yang23 exhibits a slight performance drop for Chinese text at lower scaling factors. In contrast, Wu04 suffers from severe degradation under all non-native scales, with accuracy fluctuating around 50\%, which aligns with the scale sensitivity analysis discussed in Section~\ref{strategy1}. Overall, CoreMark demonstrates the most superior performance.

\vspace{0.3em}
\noindent \textbf{Watermark Generalizability (w.r.t. Different Fonts).}
We evaluate the robustness and imperceptibility of several commonly used fonts in both Chinese and English. The font sizes for Chinese and English fonts are set to 14pt and 18pt, respectively. For each font, we test 30 text images, and Table~\ref{differentfont-robustness} records the detailed results. Under the Screenshot, CoreMark attains near-perfect accuracy, reaching up to 99.39\% on English (TNR) and 98.19\% on Chinese (DengXian). Under more challenging conditions such as Print-scan and Print-camera, CoreMark maintains robust performance, demonstrating its superior resilience to physical distortions. Yang23 performs better than Wu04, particularly on English fonts, but still falls short compared to CoreMark. These results validate the generalizability of CoreMark across diverse font styles.

\vspace{0.3em}
\noindent \textbf{Watermark Generalizability (w.r.t. Foreign Languages).}
We also evaluate the adaptability of CoreMark to languages beyond English and Chinese, including Arabic, French, Japanese, Korean, Russian, and Spanish. For each language, we test 30 text images. Table~\ref{foreign} presents the PSNR, SSIM, and $ACC$ values of six languages. The results indicate that Arabic exhibits slightly lower robustness compared to other languages, but it achieves over 90\% accuracy in print-scan and screenshot scenarios, and over 85\% accuracy in resisting print-camera. Regardless of invisibility or robustness, CoreMark performs significantly better than the baseline methods.

\begin{figure}[t]
\vspace{-0.7em}
\subfloat[]{
    \centering
    \includegraphics[width=0.105\textwidth]{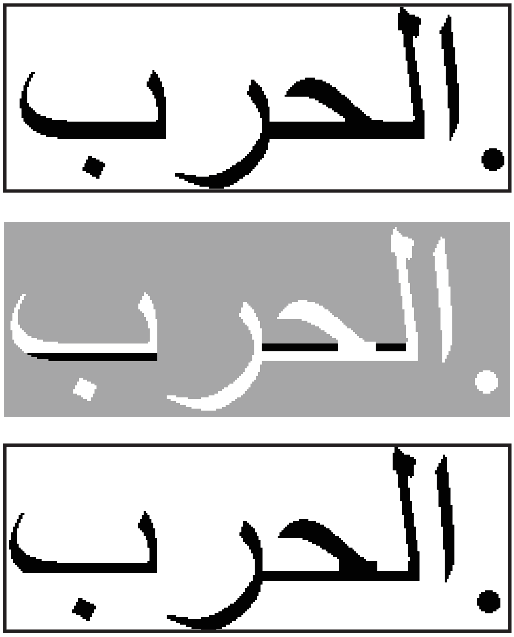}
}
\subfloat[]{
    \centering
    \includegraphics[width=0.102\textwidth]{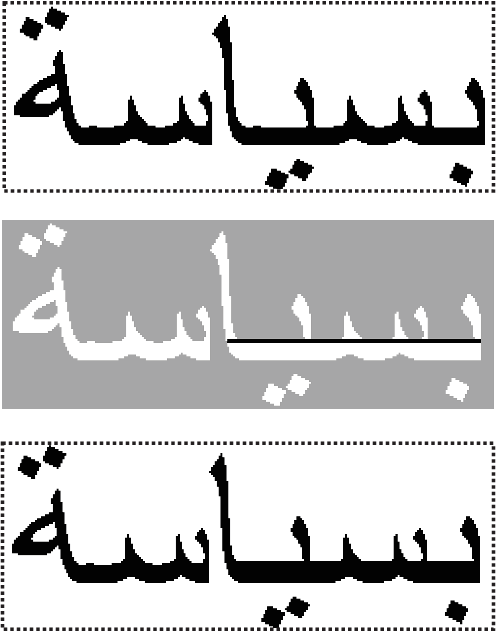}
}
\subfloat[]{
    \centering
    \includegraphics[width=0.112\textwidth]{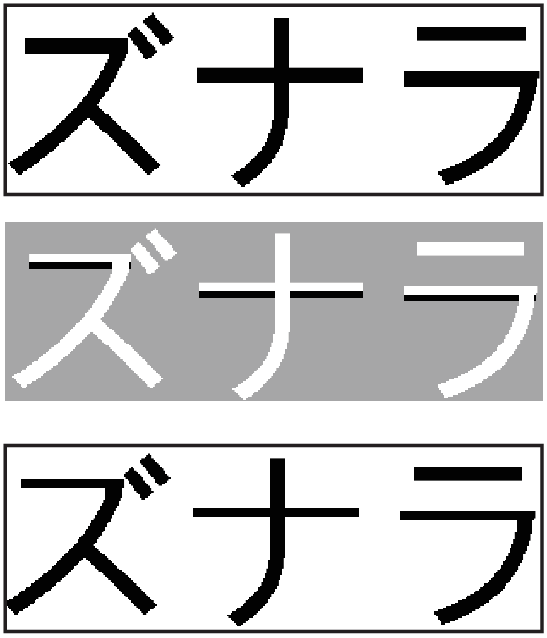}
}
\subfloat[]{
    \centering
    \includegraphics[width=0.112\textwidth]{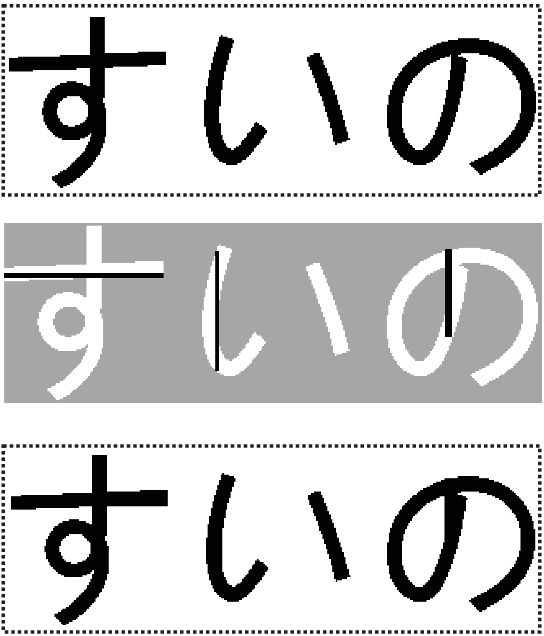}
}\newline
\subfloat[]{
    \centering
    \includegraphics[width=0.101\textwidth]{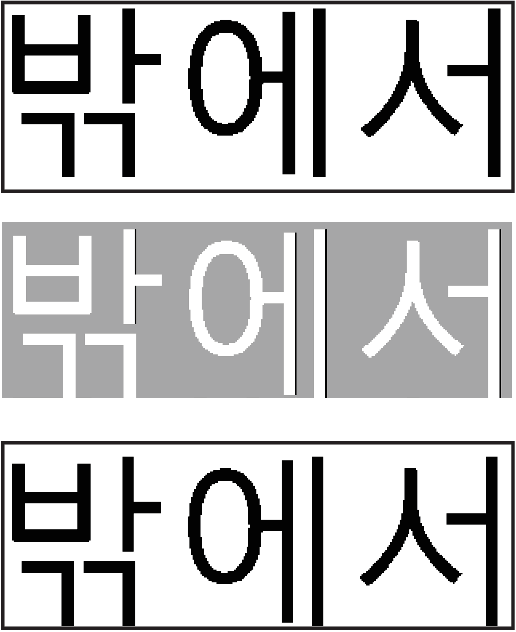}
}
\subfloat[]{
    \centering
    \includegraphics[width=0.101\textwidth]{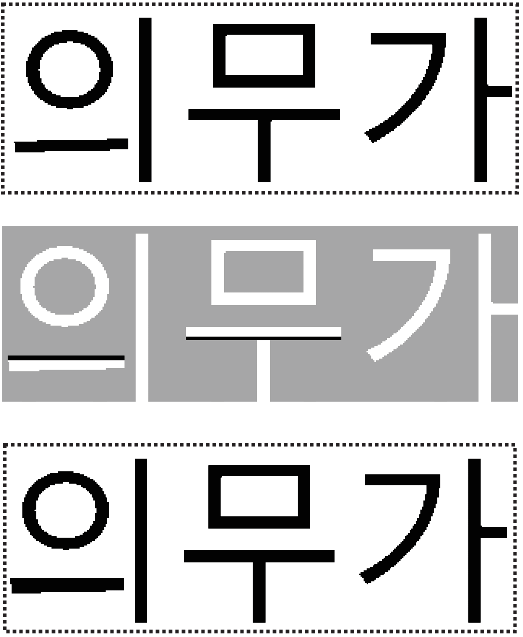}
}
\subfloat[]{
    \centering
    \includegraphics[width=0.118\textwidth]{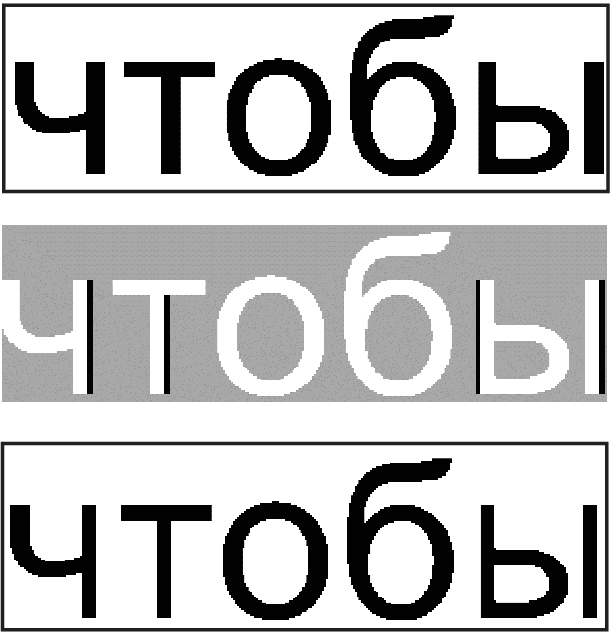}
}
\subfloat[]{
    \centering
    \includegraphics[width=0.112\textwidth]{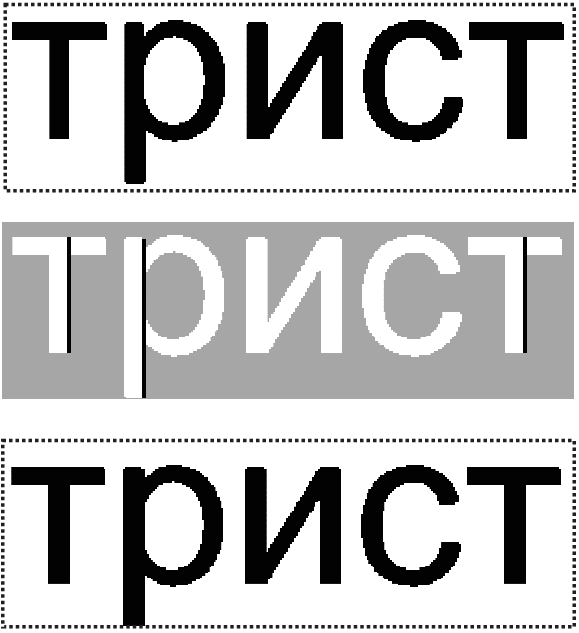}
}
\caption{Reducing examples in  (a and b) Arabic, (c) and (d) Japanese,  (e) and (f) Korean,  (g) and (h) Russian.}
\label{fig_14}
\end{figure} 

\subsection{Resistance to Potential Adaptive Attacks}
The primary objective of an attacker is to forge authentication data such that the tampered document can still pass verification tests \cite{wu2004data}. In this study, we assume that the attacker has full knowledge of the principles of CoreMark. Under this assumption, the adversary may employ two potential strategies to forge the authentication data, as detailed below.

\vspace{0.3em}
\noindent \textbf{Resistance to Key-Recovery.}
In general, the watermark information is first encrypted before being embedded into the host medium \cite{wu2004data, xiao2018fontcode}. In CoreMark, we employ BLAKE2 \cite{2014BLAKE} as a cryptographic hash function, using both the secret key and the watermark as inputs to generate an encrypted watermark. Since attackers lack knowledge of the specific parameters used, the key remains undisclosed. The probability of successfully breaking the key is $2^{\text{-len(key)}}$, indicating that longer keys provide stronger security assurances.

\vspace{0.3em}
\noindent \textbf{Resistance to Tampering.}
We further consider a more severe attack: tampering. In this scenario, the attacker may attempt to adaptively alter the watermarked carrier to remove the embedded watermark. To simulate this, we disclose the implementation details of CoreMark to three volunteers, while keeping the specific embedding parameters and watermark undisclosed. Each volunteer receives 10 text images rendered in SimSun font and is instructed to randomly modify them. We collect the modified images and attempt to extract the embedded watermarks, resulting in an $ACC$ of 83.9\%. Notably, in most cases, the watermark remains successfully extractable despite tampering, demonstrating that our method exhibits a certain level of resistance against such targeted attacks.
\begin{figure*}[t]
\subfloat[]{\includegraphics[width=0.253\textwidth]{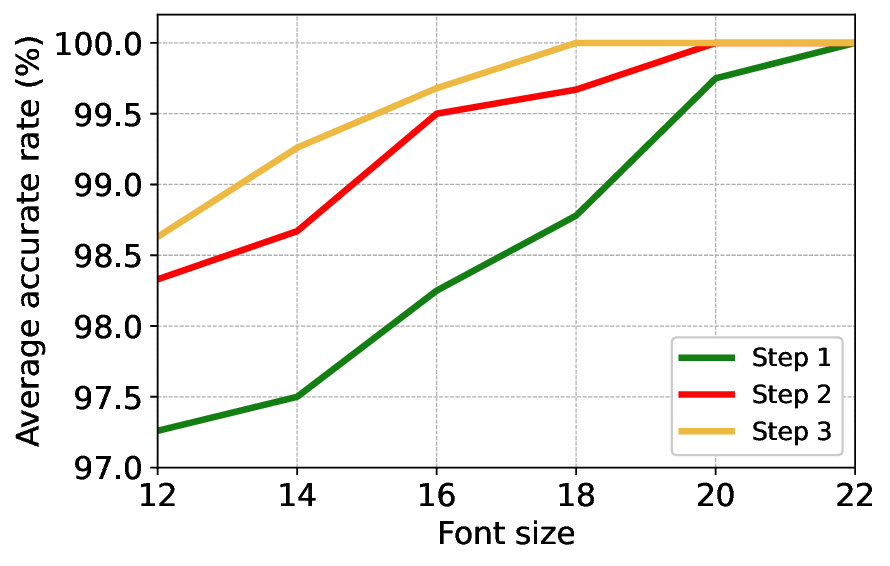}}
\subfloat[]{\includegraphics[width=0.245\textwidth]{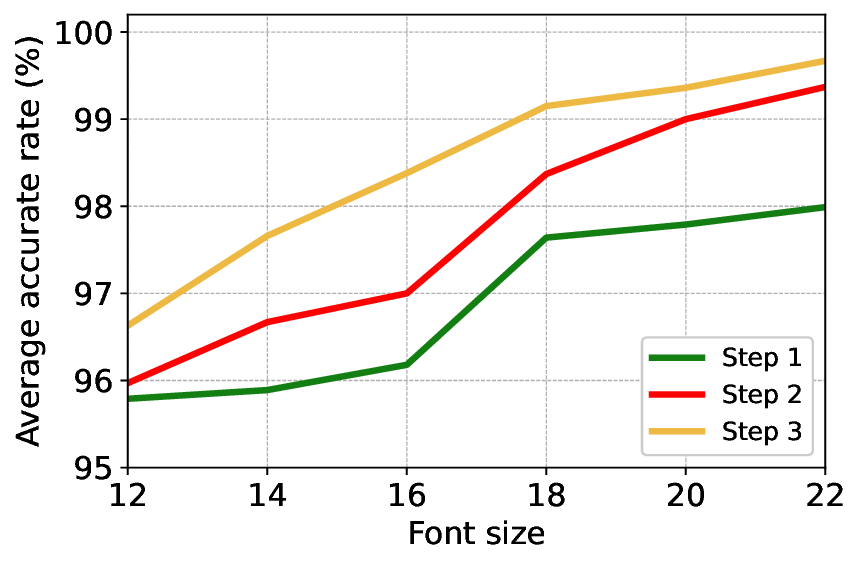}}
\subfloat[]{\includegraphics[width=0.238\textwidth]{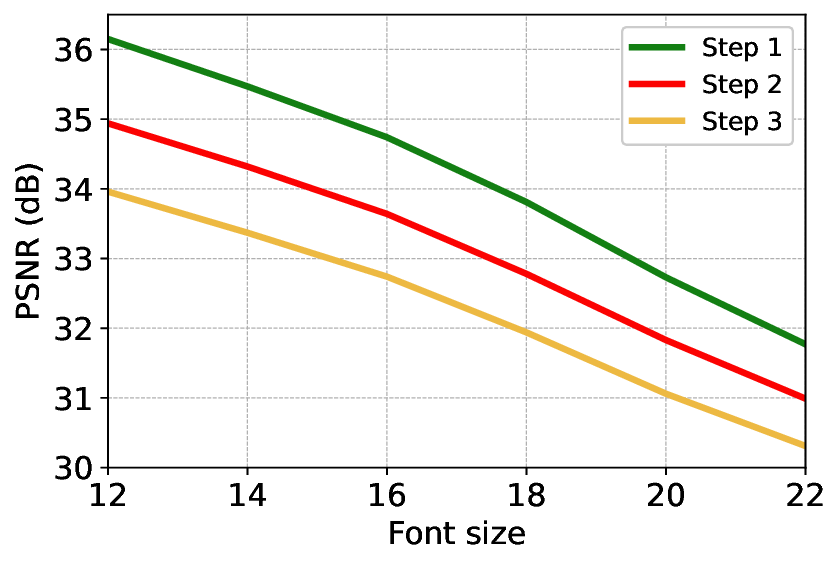}}
\subfloat[]{\includegraphics[width=0.258\textwidth]{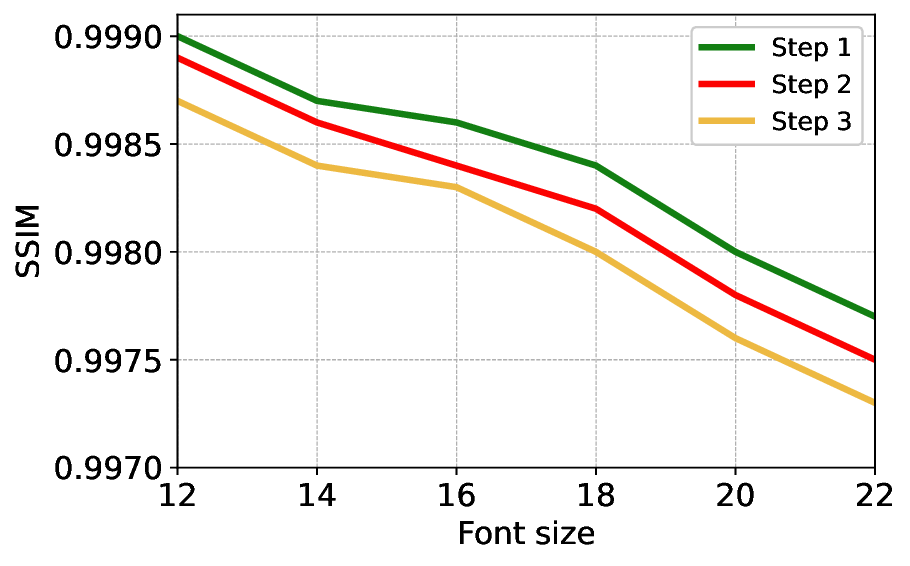}}
\centering
\caption{Ablation study of the watermark embedding strengths under different font sizes. (a) and (b) show the robustness results under screenshot and print-scan attacks respectively, while (c) and (d) present the imperceptibility metrics using PSNR and SSIM.}
\label{ablation-t}
\vspace{-0.5em}
\end{figure*}
\begin{figure}[t]
\subfloat{\includegraphics[width=0.225\textwidth]{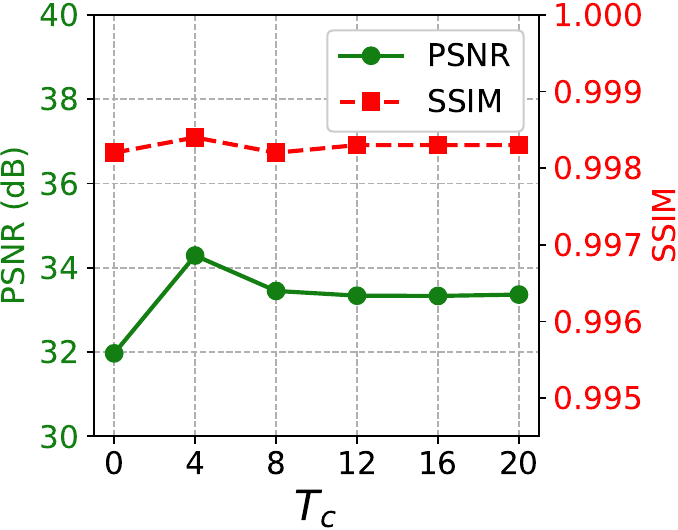}}
\hspace{0.2em}
\subfloat{\includegraphics[width=0.22\textwidth]{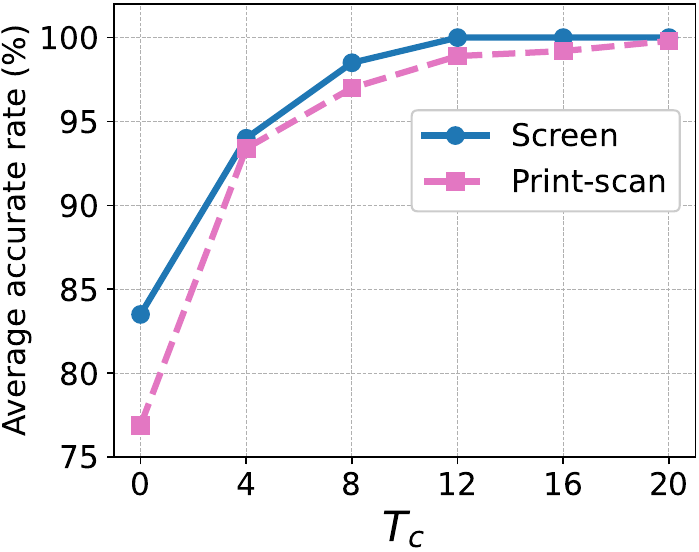}}
\centering
\caption{Effect of $T_c$ on Invisibility (left) and Robustness (right).}
\label{fig:exptc}
\vspace{-0.5em}
\end{figure}
\begin{figure}[t]
\subfloat{\includegraphics[width=0.22\textwidth]{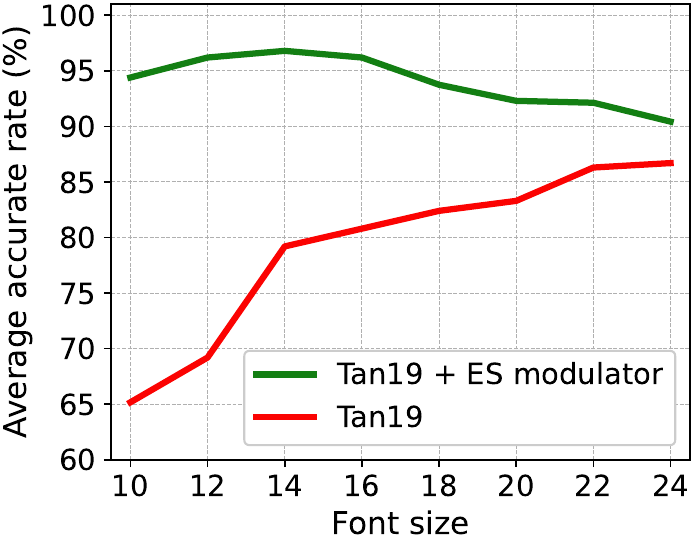}}\hspace{0.3em}
\subfloat{\includegraphics[width=0.22\textwidth]{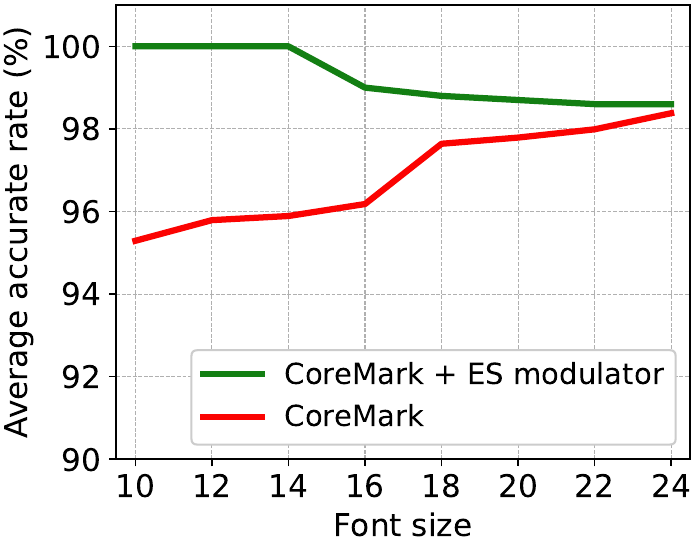}}
\centering
\caption{Average extraction accuracy under print-scanning of Tan19 (left) and CoreMark (right) with and without the ES modulator across different font sizes, when coping with print-scanning operation.}
\label{ES}
\vspace{-0.5em}
\end{figure}
\subsection{Ablation Study}

\noindent \textbf{Effects of Quantization Step $\beta$.}
In our main results, the quantization step size $\beta$ is set to 1. Since varying $\beta$ values correspond to different embedding strengths, we further investigate their impact on CoreMark's performance. Specifically, we evaluate the robustness (\eg, screenshot, print-scan) and imperceptibility of the method under three embedding strengths: $\beta = 1$, $2$, and $3$, corresponding to Step 1, 2, 3, respectively. The experiments are conducted using the Calibri font across a range of font sizes to examine the performance variations in both aspects. As illustrated in Figure~\ref{ablation-t}, increasing the embedding strength consistently enhances robustness but slightly reduces visual quality. Nevertheless, CoreMark achieves a good balance between robustness and imperceptibility. Users can flexibly select the embedding strength based on application requirements: for scenarios where visual quality is critical, Step 1 or Step 2 yields a PSNR above 33 dB and SSIM exceeding 0.998 across most font sizes; for more adversarial environments, Step 3 offers near-perfect extraction accuracy (approaching 100\%). By default, $\beta$ is set to 1. We also analyze the effect of the parameter $T_c$ in Figure~\ref{fig:exptc}, finding that while it has little impact on invisibility, overly small values may weaken robustness. Thus, setting a relatively higher value for $T_c$ is sufficient to ensure both invisibility and robustness.
\begin{figure}[t]
\subfloat{\includegraphics[width=0.22\textwidth]{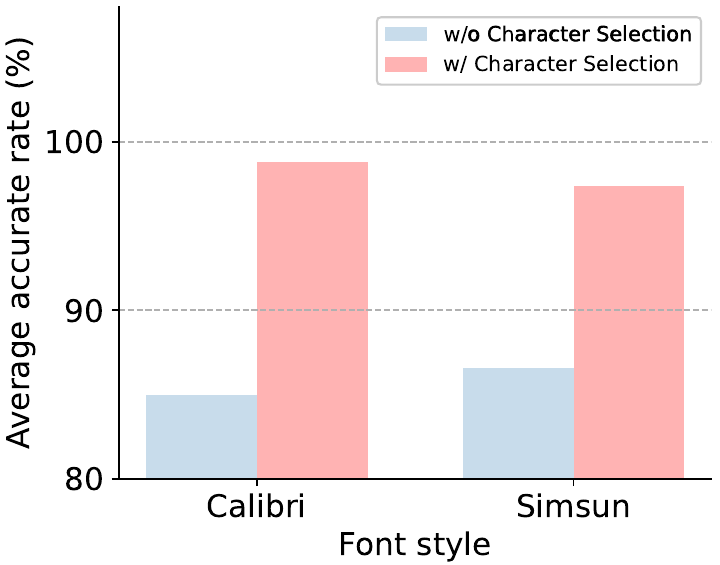}}\hspace{0.2em}
\subfloat{\includegraphics[width=0.22\textwidth]{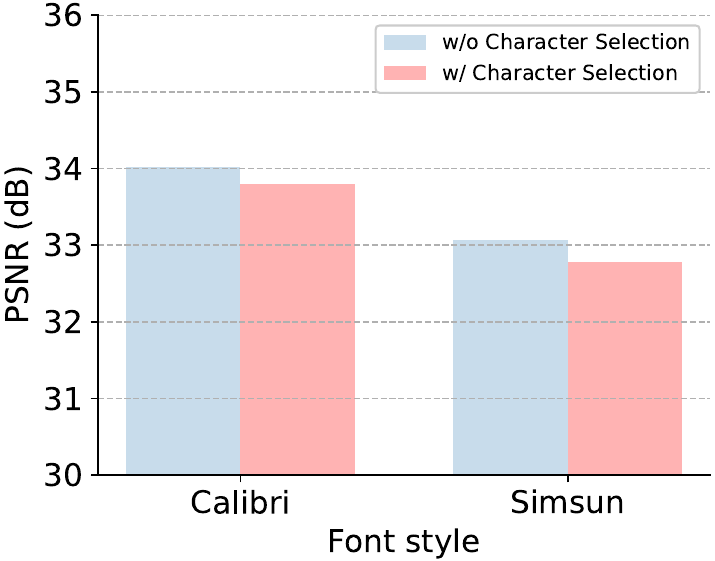}}
\centering
\caption{Effect of character selection on robustness (left) and imperceptibility (right) across different font styles (Calibri and SimSun).}
\label{fig:expselection}
\vspace{-0.5em}
\end{figure}

\vspace{0.3em}
\noindent \textbf{Effects of ES Modulator.}
To objectively assess the core performance of CoreMark, the ES modulator is excluded from all main experiments except Table~\ref{equipment}. To further verify the generality of ES modulator mentioned in Section~\ref{ES modulator}, we apply it to both our CoreMark and Tan19 \cite{tan2019print}, and evaluate its impact on robustness across different font sizes. As shown in Figure~\ref{ES}, for Tan19, the addition of the ES modulator significantly improves the extraction accuracy, especially for smaller font sizes, with the average accuracy increasing from approximately 65\% to over 95\%. For CoreMark, which already demonstrates strong robustness, the ES modulator further enhances performance, maintaining extraction accuracy close to 100\% across all font sizes. These results highlight the effectiveness and generalizability of the ES modulator in enhancing watermark robustness under challenging conditions.

\vspace{0.3em}
\noindent \textbf{Effects of Character Selection.}
Character Selection is designed to eliminate characters with low robustness, such as punctuation marks (\eg, commas), by excluding them during the embedding process. To validate its effectiveness, we conduct robustness (under print-scan distortion) and imperceptibility experiments using Calibri (English) and Simsun (Chinese) font styles. As observed in Figure~\ref{fig:expselection}, integrating character selection substantially enhances extraction accuracy for both Calibri and Simsun fonts, indicating a notable improvement in robustness. In contrast, the PSNR remains relatively stable across all settings, demonstrating that character selection has minimal impact on invisibility. These results validate the effectiveness of the character selection strategy in improving robustness without compromising perceptual quality.

\section{Potential Limitations and Future Directions}
In this section, we discuss the limitations of our method from three perspectives and outline possible directions for future research:
\textbf{1)} Unlike regular text, CoreMark encodes characters as images, rendering them uneditable and unselectable. Arguably, this limitation does not affect the core function of document protection. Besides, it can potentially be addressed by converting image characters into vector graphics, as done in AutoStegaFont \cite{yang2023autostegafont}. However, since CoreMark focuses on image-based text watermarking, this topic is beyond our scope and will be explored in future work. 
\textbf{2)} Currently, CoreMark mainly works with black and white images. Although it can be extended to colored fonts by embedding data through color-to-white pixel transitions, most documents use black text. Therefore, colored font processing is not included in this study and is left for future work.
\textbf{3)} CoreMark currently takes about 0.2 seconds per character for embedding. While this time is acceptable for practical use, we plan to optimize the algorithm in future work to further improve embedding efficiency.

\section{Conclusion}
In this paper, we analyzed existing text watermarking methods and found that they struggle to balance invisibility and robustness, with occasional embedding failure. We also revealed that methods relying on edge features are particularly vulnerable to binarization.
Based on our analysis, we propose a new embedding paradigm termed \textsc{core} to balance robustness and invisibility while maintaining generalizability across diverse languages and fonts. Building upon this paradigm, we present CoreMark, a text watermarking framework encompassing both embedding and extraction processes. The embedding process first extracts \textsc{core}s from individual characters and selects characters with superior robustness, then embeds watermarks by reducing or
expanding \textsc{core}s. The extraction process is achieved by comparing \textsc{core} widths with an adaptive threshold. Furthermore, we propose a general plug-and-play embedding strength modulator to enhance robustness for small-sized characters by adaptively adjusting the embedding strength. Experimental results show that CoreMark achieves stronger robustness with better invisibility compared with state-of-the-art methods, while maintaining satisfying generalizability. 

\label{conclusion}

\bibliographystyle{IEEEtran}
\bibliography{IEEEabrv, IEEEexample}

\end{document}